\documentclass[sigconf]{acmart}
\usepackage{multirow}
\usepackage{subfigure}
\usepackage{booktabs}
\usepackage{marvosym}
\usepackage{amsmath,epsfig}
\usepackage{url}
\usepackage{color}
\usepackage{xcolor}
\usepackage{colortbl}

\usepackage{amssymb}
\usepackage{amsfonts}
\usepackage{arydshln}
\AtBeginDocument{%
  \providecommand\BibTeX{{%
    \normalfont B\kern-0.5em{\scshape i\kern-0.25em b}\kern-0.8em\TeX}}}
\definecolor{deepred}{rgb}{0.9, 0, 0}
\definecolor{deepblue}{rgb}{0, 0, 0.8}
\newcommand{\redbold}[1]{\textbf{\textcolor{deepred}{#1}}}
\newcommand{\bluebold}[1]{\textbf{\textcolor{deepblue}{#1}}}
\copyrightyear{2024}
\acmYear{2024}
\setcopyright{acmlicensed}\acmConference[MM '24]{Proceedings of the 32nd
ACM International Conference on Multimedia}{October 28-November 1,
2024}{Melbourne, VIC, Australia}
\acmBooktitle{Proceedings of the 32nd ACM International Conference on
Multimedia (MM '24), October 28-November 1, 2024, Melbourne, VIC, Australia}
\acmDOI{10.1145/3664647.3680946}
\acmISBN{979-8-4007-0686-8/24/10}
\begin{document}

%%
%% The "title" command has an optional parameter,
%% allowing the author to define a "short title" to be used in page headers.
\title{LMM-PCQA: Assisting Point Cloud Quality Assessment with LMM}

%%
%% The "author" command and its associated commands are used to define
%% the authors and their affiliations.
%% Of note is the shared affiliation of the first two authors, and the
%% "authornote" and "authornotemark" commands
%% used to denote shared contribution to the research.
% \author{Ben Trovato}
% \authornote{Both authors contributed equally to this research.}
% \email{trovato@corporation.com}
% \orcid{1234-5678-9012}
% \author{G.K.M. Tobin}
% \authornotemark[1]
% \email{webmaster@marysville-ohio.com}
% \affiliation{%
%   \institution{Institute for Clarity in Documentation}
%   \streetaddress{P.O. Box 1212}
%   \city{Dublin}
%   \state{Ohio}
%   \country{USA}
%   \postcode{43017-6221}
% }

\author{Zicheng Zhang}
\email{zzc1998@sjtu.edu.cn}
\affiliation{
\institution{Shanghai Jiao Tong University}
\country{Shanghai, China}
}

\author{Haoning Wu}
\email{haoning001@e.ntu.edu.sg}
\affiliation{
\institution{Nanyang Technological University}
\country{Singapore}
}

\author{Yingjie Zhou}
\author{Chunyi Li}
% \email{[zyj2000,lcysyzxdxc]@sjtu.edu.cn}
\affiliation{
\institution{Shanghai Jiao Tong University}
\country{China}
}

\author{Wei Sun}
\email{sunguwei@sjtu.edu.cn}
\affiliation{
\institution{Shanghai Jiao Tong University}
\country{China}
}

\author{Chaofeng Chen}
\email{chaofeng.chen@e.ntu.edu.sg}
\affiliation{
\institution{Nanyang Technological University}
\country{Singapore}
}

\author{Xiongkuo Min$\dag$}
\authornote{Corresponding authors$\dag$.}
\email{minxiongkuo@sjtu.edu.cn}
\affiliation{
\institution{Shanghai Jiao Tong University}
\country{China}
}

\author{Xiaohong Liu}
\email{xiaohongliu@sjtu.edu.cn}
\affiliation{
\institution{Shanghai Jiao Tong University}
\country{China}
}

\author{Weisi Lin}
\email{wslin@e.ntu.edu.sg}
\affiliation{
\institution{Nanyang Technological University}
\country{Singapore}
}

\author{Guangtao Zhai$\dag$}
\authornotemark[1]
\email{zhaiguangtao@sjtu.edu.cn}
\affiliation{
\institution{Shanghai Jiao Tong University}
\country{China}
}

% \author{Zicheng Zhang$^1$, Haoning Wu$^2$, Yingjie Zhou$^1$, Chunyi Li$^1$, Wei Sun$^1$, Chaofeng Chen$^2$, \\ Xiongkuo Min$^1$, Xiaohong Liu$^1$, Weisi Lin$^2$, Guangtao Zhai$^1$ \\ }
% \affiliation{%
%   \institution{Shanghai Jiao Tong University$^1$}
%   \city{Nanyang Technological University$^2$}
% }
%% You do not have to enter your paper ID

%%
%% By default, the full list of authors will be used in the page
%% headers. Often, this list is too long, and will overlap
%% other information printed in the page headers. This command allows
%% the author to define a more concise list
%% of authors' names for this purpose.
\renewcommand{\shortauthors}{Zhang, et al.}

%%
%% The abstract is a short summary of the work to be presented in the
%% article.
\begin{abstract}
Although large multi-modality models (LMMs) have seen extensive exploration and application in various quality assessment studies, their integration into Point Cloud Quality Assessment (PCQA) remains unexplored. Given LMMs' exceptional performance and robustness in low-level vision and quality assessment tasks, this study aims to investigate the feasibility of imparting PCQA knowledge to LMMs through text supervision. To achieve this, we transform quality labels into textual descriptions during the fine-tuning phase, enabling LMMs to derive quality rating logits from 2D projections of point clouds. To compensate for the loss of perception in the 3D domain, structural features are extracted as well. These quality logits and structural features are then combined and regressed into quality scores. Our experimental results affirm the effectiveness of our approach, showcasing a novel integration of LMMs into PCQA that enhances model understanding and assessment accuracy. We hope our contributions can inspire subsequent investigations into the fusion of LMMs with PCQA, fostering advancements in 3D visual quality analysis and beyond. The code is available at https://github.com/zzc-1998/LMM-PCQA.
\end{abstract}

%%
%% The code below is generated by the tool at http://dl.acm.org/ccs.cfm.
%% Please copy and paste the code instead of the example below.
%%
\begin{CCSXML}
<ccs2012>
   <concept>
       <concept_id>10003120.10003145.10011770</concept_id>
       <concept_desc>Human-centered computing~Visualization design and evaluation methods</concept_desc>
       <concept_significance>300</concept_significance>
       </concept>
   <concept>
       <concept_id>10010147.10010178</concept_id>
       <concept_desc>Computing methodologies~Artificial intelligence</concept_desc>
       <concept_significance>500</concept_significance>
       </concept>
 </ccs2012>
\end{CCSXML}

\ccsdesc[300]{Human-centered computing~Visualization design and evaluation methods}
\ccsdesc[500]{Computing methodologies~Artificial intelligence}

%%
%% Keywords. The author(s) should pick words that accurately describe
%% the work being presented. Separate the keywords with commas.
\keywords{Large multi-modality model, Point cloud quality assessment}

%% A "teaser" image appears between the author and affiliation
%% information and the body of the document, and typically spans the
%% page.
% \begin{teaserfigure}
%   \includegraphics[width=\textwidth]{sampleteaser}
%   \caption{Seattle Mariners at Spring Training, 2010.}
%   \Description{Enjoying the baseball game from the third-base
%   seats. Ichiro Suzuki preparing to bat.}
%   \label{fig:teaser}
% \end{teaserfigure}
\begin{teaserfigure}
\centering
    \includegraphics[width=.8\linewidth]{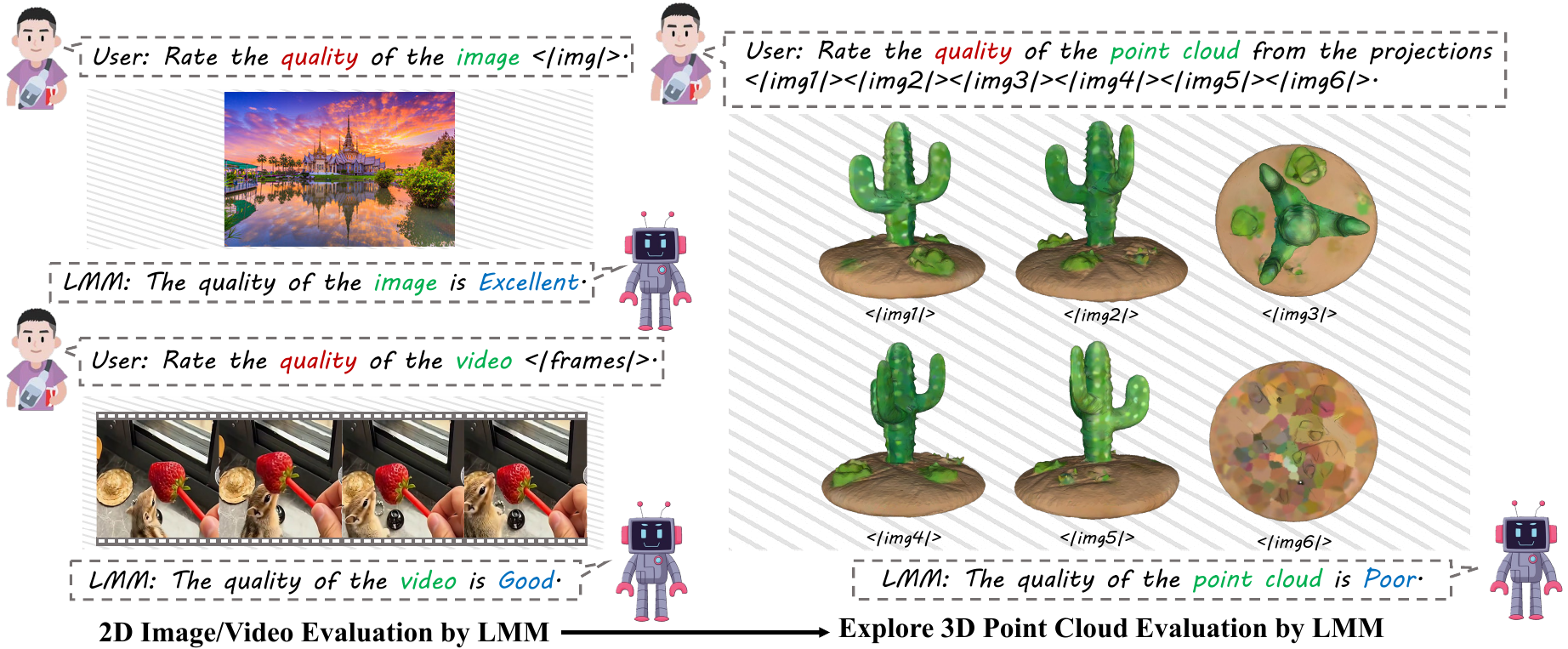} %
    \caption{Inspired by the impressive quality evaluation ability of LMM on 2D media, we are the first to explore the quality representation potential of LMM on 3D point clouds. } \label{fig:spotlight}
\end{teaserfigure}%

% \received{20 February 2007}
% \received[revised]{12 March 2009}
% \received[accepted]{5 June 2009}

%%
%% This command processes the author and affiliation and title
%% information and builds the first part of the formatted document.
\maketitle

\section{Introduction}

% Point clouds are increasingly utilized in various applications such as virtual/augmented reality \cite{park2008multiple,liu2020model,gu20193d}, autonomous vehicles \cite{cui2021deep}, and video post-production \cite{mekuria2016design}, owing to their effective 3D representation. This usage has spurred extensive research into high-level tasks like classification \cite{grilli2017review,ku2018joint,wang2019frustum,vora2020pointpainting,xie2020pi,yoo20203d,chen2020object}, detection \cite{cui2021deep}, and segmentation \cite{cheng2021sspc,liutoposeg}. In parallel, point cloud quality assessment (PCQA) has evolved to evaluate visual quality, crucial for improving simplification and compression methods \cite{fan2022d} and enhancing user experience. PCQA complexity increases with realistic point clouds due to intricate geometries and color attributes. PCQA techniques are categorized into Full-Reference (FR-PCQA), Reduced-Reference (RR-PCQA), and No-Reference (NR-PCQA), with NR-PCQA gaining prominence due to the limited availability of pristine reference clouds in practical scenarios.

Point clouds are increasingly used across diverse real-world scenarios, including virtual/augmented reality \cite{park2008multiple,liu2020model,gu20193d}, autonomous vehicles \cite{cui2021deep}, and video post-production \cite{mekuria2016design}. This surge is attributed to their adeptness in three-dimensional representation. Consequently, significant research efforts have been channeled towards enhancing high-level areas like point cloud classification \cite{grilli2017review,ku2018joint,wang2019frustum,vora2020pointpainting,xie2020pi,yoo20203d,chen2020object}, detection \cite{cui2021deep}, and segmentation \cite{cheng2021sspc,liutoposeg}. Meanwhile, as a key component for ensuring point cloud quality, point cloud quality assessment (PCQA) has seen comparable advancements as well during the last decade. {PCQA's objective is to appraise point clouds' visual quality}, a pivotal factor for refining simplification and compression strategies in practical applications \cite{fan2022d}, and to elevate the Quality of Experience (QoE) for end-users. Generally, PCQA methods are divided into three categories, depending on their reliance on reference point clouds: Full-Reference PCQA (FR-PCQA), Reduced-Reference PCQA (RR-PCQA), and No-Reference PCQA (NR-PCQA) respective. The availability of pristine reference point clouds is often limited in real-world applications, {emphasizing the demand for NR-PCQA approaches}, which is why our research primarily focuses on NR-PCQA.

% The complexity of PCQA is further compounded when assessing point clouds depicting realistic objects or individuals, due to their intricate geometric structures and the dense aggregation of points, often augmented with color attributes, posing additional challenges. 
% \begin{figure}[t]
%     \centering
%     \includegraphics[width=\linewidth]{figs/spotlight.pdf}
%     \caption{An example of prompting LMM for PCQA, where point cloud can be regarded as a sequence of projections and \textit{[SCORE\_TOKEN]} indicates the quality prediction score token responded by LMM.}
%     \label{fig:spotlight}
%     \vspace{-0.5cm}
% \end{figure}

There are already some cutting-edge studies that have begun applying Large Multi-Modality Models (LMMs) to low-level vision and quality assessment fields \cite{wu2023qbench,wu2023qinstruct,wu2023qalign,depictqa,visualcritic}, achieving notable success. LMMs demonstrate highly competitive performance and robustness in these tasks. However, the main focus of these studies has remained on two-dimensional (2D) media such as images and videos, \textbf{while no research has explored the possibility of applying LMMs to three-dimensional (3D) media like point clouds.} It is known that both 2D and 3D media exhibit similar distortions, i.e., blur and noise. Given the robust quality perception of LMMs in 2D, \textbf{we can hold the hypothesis that LMM also has significant quality perception abilities in 3D point clouds.} Hence, investigating the application of LMMs for point cloud quality assessment is not only valuable but also meaningful, reflecting their established visual perception strengths. Therefore, in this study, we carry out a novel method named \textbf{LMM-PCQA} to provide an interesting solution for handling the PCQA problem with the assistance of LMM.

First, \textbf{we treat the point clouds as sequences of projections} to enable LMM to perceive the point cloud visual quality. Afterward, we try to \textbf{teach LMM about the quality alignment} between the predefined 5-level \textbf{qualitative adjectives} (i.e., \textit{excellent, good, fair, poor, bad}) and point cloud projections. Specifically, we employ the existing PCQA databases to provide the necessary knowledge, where the quality labels are transformed into corresponding \textbf{qualitative adjectives}. Then we specially design a prompt structure to produce the question-answer pairs, which are composed of the question \textit{`Rate the quality of the point cloud from the projections [img1],[img2],[img3],[img4],[img5],[img6]'} and the answer \textit{`The quality of the point cloud is excellent/good/fair/poor/bad'}. These question-answer pairs can be utilized to teach LMM PCQA knowledge during the fine-tuning stage. After the instruction tuning, we can expect the trained LMM to give the predicted \textit{[SCORE\_TOKEN]} with the same prompt structure (the \textbf{qualitative adjectives} position is left blank for LMM response), which is shown in Fig. \ref{fig:framework}. The predicted \textit{[SCORE\_TOKEN]} can be recognized as a probability map to the \textbf{qualitative adjectives}, and we convert it into 5-level probabilities as the LMM evaluation results.

Secondly, to address the potential insensitivity to geometric distortions (e.g., compression, downsampling) when only analyzing projections, \textbf{we propose the extraction of multi-scale structural features.} This approach enhances the LMM-PCQA's holistic comprehension of point cloud visual quality. The point clouds are converted into quality-aware structural domains, a technique validated for effective quality feature extraction in prior research \cite{alexiou2021pointpca,zhou2023pointpca+,zhang2021no}. We modify the scale parameters in the k-nearest neighbors (k-NN) algorithm to offer a multi-scale perspective, aligning with the human vision system's perception mechanism. Subsequently, key statistical parameters are used to quantify structural distortions within these domains. Finally, we combine the LMM's evaluative results and the structural features, utilizing support vector regression (SVR) to derive the quality values. The experimental outcomes affirm that our LMM-PCQA model is on par with, or surpasses, current leading PCQA methods.

By carrying out LMM-PCQA, the contributions of this paper can be summarized as follows:
\begin{itemize}
    \item \textbf{We are the first to employ LMM for PCQA tasks.} We design a novel prompt structure to enable the LMM to perceive the point cloud visual quality. The existing PCQA databases are converted into question-answer pairs, which are then used to \textbf{inject PCQA knowledge into LMM}. 
    \item  \textbf{We propose the extraction of multi-scale structural features.} By processing point clouds into multi-scale domains, we quantify the geometry distortions via key statistic parameters estimation, which helps LMM gain a more comprehensive understanding of point cloud visual quality.
    \item  \textbf{LMM-PCQA demonstrates exceptional performance across various PCQA databases.} The ablation study and cross-database evaluations further validate the logical design of LMM-PCQA and its robust generalization capabilities.
\end{itemize}

\section{Related Works}

\subsection{LMM for Quality Assessment}
Recent studies are exploring LMMs in visual quality assessment. X-IQE~\cite{chen2023x} uses them to evaluate text-to-image generation via a chain of thoughts strategy. Q-Bench~\cite{wu2023qbench} introduces a binary softmax method for LMMs to produce quantifiable scores by assessing tokens \textit{good} and \textit{poor}. Q-Instruct~\cite{wu2023qinstruct} improves this by fine-tuning LMMs through text-based questioning for specific visual queries. Q-Align~\cite{wu2023qalign} adopts human-like evaluation mechanisms, enhancing visual quality scoring.

\begin{figure*}
    \centering
    \includegraphics[width=.85\linewidth]{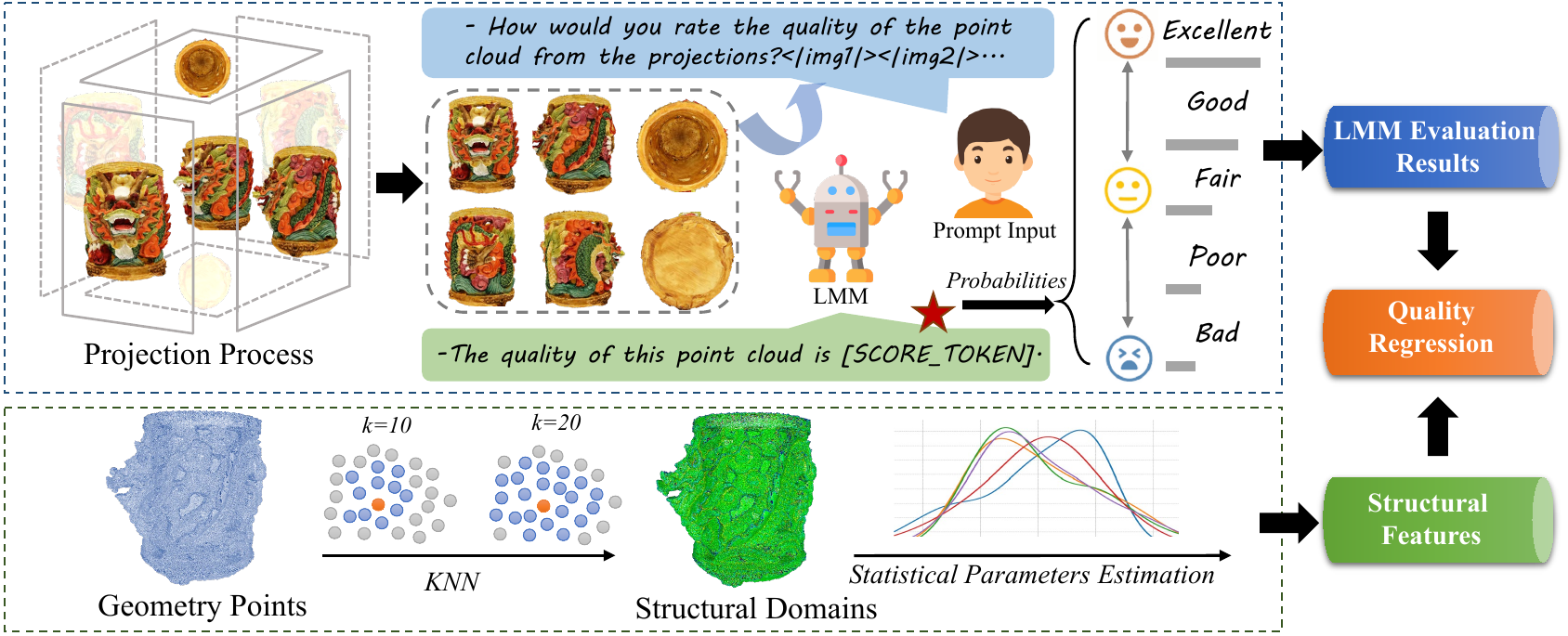} 
    \caption{The framework of the proposed method.}
    \label{fig:framework}
    \vspace{-0.4cm}
\end{figure*}

\subsection{PCQA Development}
In the early stages of PCQA, the MPEG group developed Full-Reference PCQA (FR-PCQA) methods like p2point \cite{mekuria2016evaluation} and p2plane \cite{tian2017geometric} for point cloud quality assessment, and later introduced a color-based PSNR-yuv method \cite{torlig2018novel}. Despite challenges in addressing complex distortions, advanced FR-PCQA metrics like PCQM \cite{meynet2020pcqm}, GraphSIM \cite{yang2020graphsim}, and PointSSIM \cite{alexiou2020pointssim} were developed to integrate structural features for enhanced performance.

Building on progress in no-reference image and video quality assessment (NR-I/VQA) \cite{zhang2015application,liu2015paraboost,gu2017learning,zhang2022nor,zhang2022dual,zhang2021ano}, various NR-PCQA methods have evolved. Techniques include using CNNs for quality regression with handcrafted features \cite{chetouani2021deep}, multi-view projections for feature extraction \cite{liu2021pqa,zhang2021no}, and direct prediction with sparse CNNs \cite{liu2022point}. Innovative approaches also involve converting point clouds to videos for VQA \cite{fan2022no,zhang2022treating}, employing structure-guided resampling \cite{zhou2022blind}, integrating multi-projections for simpler feature extraction \cite{zhang2023gms}, adapting image quality assessment methods for point cloud rendering \cite{yang2022no}, and using non-local geometry and color gradient models for quality estimation \cite{wang2023non}. Recent frameworks like MM-PCQA \cite{zhang2022mm} and pmBQA \cite{xie2023pmbqa} leverage multi-modal learning for improved PCQA. Emerging research explores text modality's potential in quality assessment learning.

\section{Proposed Method}
The framework of the proposed method is briefly illustrated in Fig. \ref{fig:framework}, which includes the LMM evaluation module, the structural feature extraction module, and the quality regression module.

\subsection{LMM Evaluation}

\subsubsection{Projection Acquisition}
Consider a colored point cloud $\mathbf{P} = {(p_{i}, c_{i})}_{i=1}^{N}$, where $p_{i} \in \mathbb{R}^{1 \times 3}$ represents the single point consisting of geometry coordinates, $c_{i} \in \mathbb{R}^{1 \times 3}$ represents the RGB color attributes, and $N$ denotes the total count of points. We then adopt the conventional cube-like viewpoints configuration, which is inspired by the projections approach discussed in \cite{torlig2018novel,alexiou2019exploiting}. As illustrated in Fig. \ref{fig:framework}, six orthogonal viewpoints are utilized, each mapping to one of the cube's six faces for generating the projections. For a point cloud $\mathbf{P}$, the rendering process is derived as:

\begin{equation}
\begin{aligned}
     &\mathbf{I}  = \psi(\mathbf{P}), \\
    \mathbf{I}  = \{&\mathcal{I}_{k}|k =1,\cdots, 6\},
\end{aligned}
\end{equation}
where $\mathbf{I}$ represents the set of the 6 rendered projections and $\psi(\cdot)$ denotes the Open3D-based \cite{zhou2018open3d} projections capturing process. 

\subsubsection{How to inject PCQA knowledge into LMMs ?} LMM has shown competitive performance in 2D image/video quality assessment \cite{wu2023qalign,wu2023qbench,wu2023qinstruct}, therefore it is feasible to apply LMM for PCQA tasks by taking the point cloud as a sequence of projections. Then it is vital to solve the core problem of \textbf{PCQA Knowledge Injection}.
Following the common approaches of training LMMs \cite{mplug,llava}, it is natural to come up with the solution of instructing the LMM with question-answer pairs regarding PCQA issues. Thus we carry out the specific prompt structure as follows:
\newline

    \textit{-How would you rate the quality of the point cloud from the projections?<|img1|><|img2|>...}
    
    \textit{-The quality of this point cloud is [QA(mos)].}
\newline

\noindent where \textit{<|img1|><|img2|>...} stands for the image set of projections and \textit{QA(s)} is the \textbf{qualitative adjective} of the point cloud which can be obtained from the mean opinion score (MOS) corresponding to the point cloud. Afterward, we can use the designed question-answer pairs to teach LMM PCQA knowledge.

\subsubsection{Transformation from MOS to quality rating.} 
In daily experiences, humans often give feedback using \textbf{qualitative adjectives} \textit{(such as good, bad, superb)} instead of \textbf{numerical ratings} \textit{(like 9.2, 2.5, 7.1)}. Hence, implementing visual scoring activities with level-based ratings taps into this \textit{natural tendency} of humans (\textit{to offer qualitative adjectives}). 
Similarly, the perception and expression of LMM are akin to humans, which have a better understanding and perception of \textbf{qualitative adjectives}. Therefore, converting MOS into corresponding \textbf{qualitative adjectives} for its learning is more intuitive than having it learn directly from numbers. 
Specifically, the transformation of MOS can be achieved by evenly splitting the range from the highest score ($\mathrm{M}$) to the lowest score ($\mathrm{m}$) into five unique intervals, with scores in each interval designated as corresponding quality levels:

\begin{equation}
    {QA(mos)} \!=\!w_i \text{  if } \text{m} + \frac{i-1}{5}  \times \mathrm{(M-m)} \!<\! mos \!\leq\! \mathrm{m} + \frac{i}{5} \times \mathrm{(M-m)},
\end{equation}
where \{$w_i|_{i=1}^{5}\}=\{\textit{bad, poor, fair, good, excellent}\}$ are the standard text rating levels as defined by ITU~\cite{itu}.

\begin{figure}
    \centering
    \includegraphics[width=\linewidth]{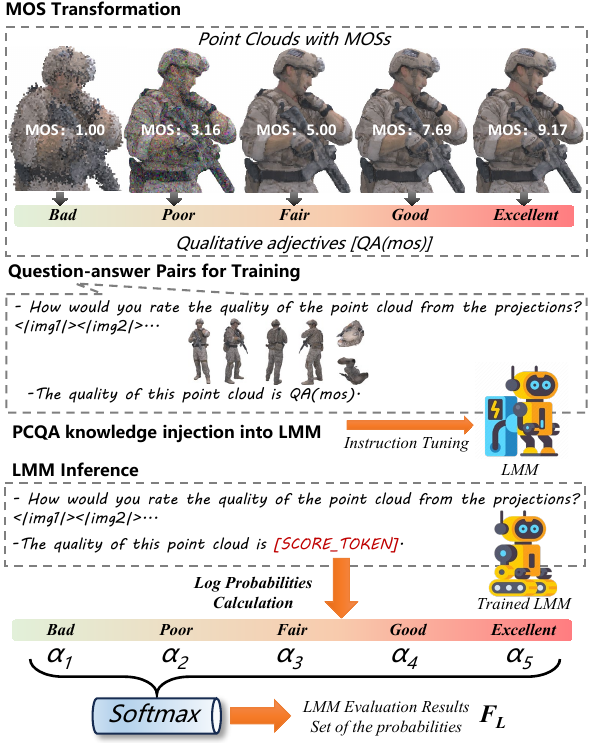}
    \caption{Illustration of the LLM evaluation pipeline. The point clouds with MOSs are transformed into question-answer pairs for LMM tuning. The LMM evaluation results can be obtained as the set of the probabilities to the predefined \textbf{qualitative adjectives}. }
    \label{fig:llm_evaluation}
\end{figure}

\subsubsection{Obtaining evaluation results via LMM inference.}
After training, we can get the evaluation results with the same prompt structure and get the response \textit{[SCORE\_TOKEN]} via LMM inference. The \textit{[SCORE\_TOKEN]} can be recognized as a log probability map to the \textbf{qualitative adjectives}. Then we can compute the final probabilities to the 5-level \textbf{qualitative adjectives} from the corresponding log probabilities via softmax as the LMM evaluation results:

\begin{equation}
    F_{L} = \{\frac{e^{\alpha_i}}{\sum_{j=1}^5 e^{\alpha_j}}\}_{i=1}^5,
\end{equation}
where $\alpha_i$ indicates the log probability of $i$-th \textbf{qualitative adjectives} and $F_L$ represents the LMM evaluation results which consist of 5 probabilities after softmax.

\subsection{Structual Feature Extraction}
The inadequacy of single-modality information from projections for point cloud quality evaluation has been established \cite{zhang2022mm,xie2023pmbqa}. Therefore, our approach enhances the accuracy of LMM evaluation results (projections only) by integrating geometric structural features, aiming for a more detailed and accurate assessment.

% It is widely recognized that relying only on single-modality information derived from projections is insufficient for a comprehensive evaluation of point cloud visual quality \cite{zhang2022mm,xie2023pmbqa}. Consequently, our approach involves extracting structural features from the geometric aspect to enhance the model's understanding of point cloud quality, aiming for a more thorough and precise assessment.

\subsubsection{Structural Domain}
Given the point cloud $\mathbf{P} = {(p_{i}, c_{i})}_{i=1}^{N}$, the neighborhood $P_{Nbi}$ of each point $p_{i}$ can be obtained utilizing the k-nearest neighbors (k-NN) algorithm:
\begin{equation}
\begin{aligned}
    & \quad \quad \mathbf{P_{Nb}} = \mathop{\boldsymbol{\rm{KNN}}}\limits(\mathbf{P}),\\
   Dist(p,q)\! &=\! \sqrt{\left(p_{x}\!-\!q_{x}\right)^{2}\!+\!\left(p_{y}\!-\!q_{y}\right)^{2}\!+\!\left(p_{z}\!-\!q_{z}\right)^{2}},
\end{aligned}
\end{equation}  
where $N$ represents the total count of points within the point cloud, the term $\mathbf{P_{Nb}}$ denotes the collection of neighborhoods, while $\boldsymbol{\rm{KNN}}(\cdot)$ signifies the function of the k-nearest neighbors algorithm. The distance between points $p$ and $q$ is calculated using the Euclidean distance, expressed as $Dist(p,q)$. Given the neighborhood set $P_{Nbi}$ of point $p_i$, we can define the covariance matrix $C_{i}$ for each point $p_{i}$, which is characterized by its 3D geometric coordinates:
\begin{equation}
\begin{aligned}
     C_{i} = &  \frac{1}{K} \sum_{j=1}^{K} ({p}_{n_{j}}-\hat{p})({p}_{n_{j}}-\hat{p})^{\top},\\
    &  \{{p}_{n_{1}}, \cdots ,{p_{n_{K}}} \} \in P_{Nbi},
\end{aligned}
\end{equation}
where the term $K$ denotes size of neighborhood set $P_{Nbi}$, ${p}_{n_{j}}$ is the $j$-th neighboring point in $P_{Nbi}$, $\hat{p}$ is the centroid of this neighborhood, ${p}_{n_{j}}$ and $\hat{p}$ are vectors with dimensions $\mathbb{R}^{3 \times 1}$, while $C_{i}$ is a matrix with dimensions $\mathbb{R}^{3 \times 3}$. Consequently, the eigenvectors for the covariance matrix $C_{i}$ can be derived as follows:
\begin{equation}
C_{i} \cdot {v}_{l}=\lambda_{l} \cdot {v}_{l}, l \in\{1,2,3\},
\end{equation}
where ($\lambda_{1}$, $\lambda_{2}$, $\lambda_{3}$) stand for the eigenvalues and (${v}{1}$, ${v}{2}$, ${v}{3}$) represent the respective eigenvectors, with $\lambda{1}>\lambda_{2}>\lambda_{3}$. Consequently, we derive three eigenvalues for each point $p_{i}$ within the point cloud $\mathbf{P}$. Then we can compute the \textit{linearity} and \textit{planarity} structural domains of the point cloud as:
\begin{equation}
\begin{aligned}
    {Lin(p_{i})}&= \frac{\lambda_{1}-\lambda_{2}}{\lambda_{1}}, \\
    {Pla(p_{i})}&= \frac{\lambda_{2}-\lambda_{3}}{\lambda_{1}}, 
\end{aligned}
\end{equation}
where $Lin(p_{i})$ and $Pla(p_{i})$ represent the \textit{linearity} and \textit{planarity} values for point $p_i$. The chosen structural domains (\textit{linearity}, \textit{planarity}) have been demonstrated to exhibit a strong correlation with geometric visual losses, such as compression and downsampling, and have been extensively utilized in numerous PCQA tasks \cite{alexiou2021pointpca,zhou2023pointpca+,zhang2021no}.

\subsubsection{Multi-scale Perception} The multi-scale nature of point cloud visual perception has been noted in the literature \cite{zhang2021ms}. To account for this, we compute structural domains across various scales by varying the scale parameter $k$ in the $\boldsymbol{\rm{KNN}}(\cdot)$ process. This approach allows us to derive the multi-scale structural domains as follows:
\begin{equation}
    \mathcal{D}_{k=k_{ms}} = \mathcal{S}(\mathop{\boldsymbol{\rm{KNN}}}\limits_{k=k_{ms}}(\mathbf{P})),  \mathcal{D} \in \{Lin, Pla\},
\end{equation}
where $\mathcal{D}_{k=k_{ms}}$ denotes the multi-scale structural domains, with $k_{ms}$ represents the set of scale parameters, and $\mathcal{S}(\cdot)$ refers to the process of calculating structural domains as previously described. In our study, we establish the default set of scale parameters as $\{10,20\}$, signifying that the \textit{linearity} and \textit{planarity} domains are computed using the 10-nearest-neighbor and 20-nearest-neighbor configurations, respectively.

\subsubsection{Statistical Parameters Estimation}
To quantify the quality representation from the structural domains, we employ some basic statistical parameters estimation process:
\begin{equation}
\begin{aligned}
    F_S = \{avg(\mathcal{D}_{k=k_{ms}}), & std(\mathcal{D}_{k=k_{ms}}), ent(\mathcal{D}_{k=k_{ms}})\}, \\
    & \mathcal{D} \in \{Lin, Pla\},
\end{aligned}  
\end{equation}
where $avg(\cdot)$, $std(\cdot)$, and $ent(\cdot)$ represent the average function, standard deviation function, and entropy function respectively, and $F_S$ indicates the set of the final extracted structural features.

\subsection{Quality Regression}
To clearly demonstrate the efficacy of the proposed features, we merge the LMM evaluation results with the structural features, and then incorporate them into the visual quality score using support vector regression (SVR):
\begin{equation}
    Q = \mathbf{SVR}(F_L \oplus F_S),
\end{equation}
where $Q$ indicates the quality values, $\mathbf{SVR}(\cdot)$ represents the SVR regression process, and $\oplus$ denotes the concatenation process. 

\section{Experiment}
\subsection{Validation Databases}
To assess the efficacy of the proposed method, we employ the SJTU-PCQA database \cite{yang2020predicting}, the Waterloo point cloud assessment database (WPC)\cite{liu2022perceptual}, and the WPC2.0 database \cite{liu2021reduced} for validation. The SJTU-PCQA database contains 9 reference point clouds, subjected to seven distortion types (compression, color noise, geometric shift, down-sampling, and three mixed distortions) at six levels, yielding a total of 378 ($9\times7\times6$) distorted point clouds. The WPC database includes 20 reference point clouds, each modified by four distortions (down-sampling, Gaussian white noise, Geometry-based Point Cloud Compression (G-PCC), and Video-based Point Cloud Compression (V-PCC)), resulting in 740 ($20\times37$) distorted point clouds. Meanwhile, the WPC2.0 database features 16 reference point clouds, each undergoing 25 different V-PCC degradation settings, leading to 400 ($16\times25$) distorted point clouds.

\begin{table*}[th]
\centering
\renewcommand\tabcolsep{3.3pt} 
\renewcommand{\arraystretch}{1.0}
\caption{Performance on the SJTU-PCQA, WPC, and WPC2.0 databases. Best in \redbold{red}, second in \textcolor{deepblue}{blue}. }
\begin{tabular}{c:l|cccc|cccc|cccc}
\toprule
\multirow{2}{*}{Type}  &\multirow{2}{*}{Methods}  & \multicolumn{4}{c|}{SJTU-PCQA} & \multicolumn{4}{c|}{WPC} & \multicolumn{4}{c}{WPC2.0}\\ 
\cline{3-14}
         & & SRCC$\uparrow$       & PLCC$\uparrow$      & KRCC$\uparrow$     & RMSE$\downarrow$     & SRCC$\uparrow$      & PLCC$\uparrow$      & KRCC$\uparrow$       & RMSE$\downarrow$  & SRCC$\uparrow$      & PLCC$\uparrow$      & KRCC$\uparrow$       & RMSE$\downarrow$ \\ \hline
\multirow{8}{*}{FR}
 & MSE-p2po  & 0.7294 & 0.8123 & 0.5617 & 1.3613 & 0.4558 & 0.4852 & 0.3182 & 19.8943 & 0.4315 & 0.4626 & 0.3082 & 19.1605           \\
 & HD-p2po  & 0.7157 & 0.7753 & 0.5447 & 1.4475 &0.2786 &0.3972&0.1943 &20.8990 &0.3587 & 0.4561 & 0.2641 & 18.8976           \\
 & MSE-p2pl & 0.6277 & 0.5940 & 0.4825 & 2.2815 & 0.3281 & 0.2695 &0.2249 & 22.8226 & 0.4136 & 0.4104 & 0.2965 & 21.0400           \\
 & HD-p2pl  & 0.6441   & 0.6874    & 0.4565    & 2.1255 & 0.2827 & 0.2753 &0.1696 &21.9893  & 0.4074 & 0.4402 & 0.3174 & 19.5154  \\
 & PSNR-yuv  & 0.7950 & 0.8170 & 0.6196 & 1.3151 & 0.4493 & 0.5304 & 0.3198 & 19.3119 & 0.3732 & 0.3557 & 0.2277 & 20.1465\\
 & PCQM     & {0.8644}   & {0.8853}    & {0.7086}     & {1.0862}     &{0.7434}    & {0.7499}   & {0.5601}   & 15.1639    &0.6825 & 0.6923 & 0.4929 & 15.6314            \\
 & GraphSIM  & {0.8783}    & {0.8449}    & {0.6947}   & {1.0321}  & 0.5831    & 0.6163    & 0.4194   & 17.1939 & 0.7405 & {0.7512} & {0.5533} & {14.9922}\\
 & PointSSIM    & 0.6867  & 0.7136  & 0.4964 & 1.7001  & 0.4542    & 0.4667    & 0.3278   & 20.2733  & 0.4810 & 0.4705 & 0.2978 & 19.3917 \\ \hdashline
%  \multirow{1}{*}{RR} &
% P & PCMRR     & 0.47  & 0.68  & 0.33 & 1.72 & 0.31 & 0.33 & 0.22 & 21.42  \\ \hdashline
 \multirow{10}{*}{NR} 
 &BRISQUE  & 0.3975    & 0.4214  & 0.2966 & 2.0937  & 0.2614    & 0.3155  & 0.2088 & 21.1736  & 0.0820  & 0.3353	& 0.0487 & 21.6679
\\
 &NIQE  &0.1379 &0.2420 &0.1009 &2.2622  &0.1136 & 0.2225 &0.0953 &23.1415 & 0.1865 & 0.2925 & 0.1335 & 22.5146
 \\
 &IL-NIQE  & 0.0837 & 0.1603 & 0.0594 &2.3378 &0.0913 &0.1422 & 0.0853 &24.0133 & 0.0911 & 0.1233 & 0.0714 & 23.9987\\
 &IT-PCQA &0.8651 & 0.8283 & 0.6430 & 1.1661 & 0.4870 & 0.4329 & 0.3006 & 19.8960 & 0.5661 & 0.5432 & 0.3477 & 18.7224 \\
 &ResSCNN  & 0.8600 & 0.8100 & - & - & - & - & - & - & {0.7500} & 0.7200 &-&- \\
 &PQA-net  & 0.8372   & {0.8586}    & 0.6304 & {1.0719}  & {0.7026}    & {0.7122}    & {0.4939}   & {15.0812} & 0.6191 & 0.6426 & 0.4606 & 16.9756    \\
 &3D-NSS     & 0.7144 & 0.7382  & 0.5174 & 1.7686    & 0.6479    & 0.6514    & 0.4417   & 16.5716 & 0.5077 & 0.5699 & 0.3638 & 17.7219 \\
 &GMS-3DQA & \bluebold{0.9108} & 0.9177 & 0.7735 & 0.7872 & 0.8308 & 0.8338 & 0.6457 & \bluebold{12.2292} & \bluebold{0.8272} & \bluebold{0.8218} & \bluebold{0.6277} & \bluebold{12.9904} \\
 &{MM-PCQA}     & {0.9103}   & \bluebold{0.9226}   & \bluebold{0.7838} & \bluebold{0.7716} & \bluebold{0.8414}    & \bluebold{0.8556}    & \bluebold{0.6513}  &{12.3506} &{0.8023}    &{0.8024}    &{0.6202}  & {13.4289} \\ 
 % &Plain-PCQA & 0.9302 & 0.9133 & 0.7603 & 0.8607 & 0.8783 & 0.8793 & 0.6951 & 12.3506 & - & - & - & - \\
 % \rowcolor{gray!25} 
  &\textbf{LMM-PCQA(Ours)}     & \redbold{0.9376}  & \redbold{0.9404}   & \redbold{0.8002} & \redbold{0.7175} & \redbold{0.8825}    & \redbold{0.8739}    & \redbold{0.7064}  & \redbold{11.8171} &\redbold{0.8614}    & \redbold{0.8634}    & \redbold{0.6723}  & \redbold{10.6924} \\

                      \bottomrule
\end{tabular}
\label{tab:experiment}
\vspace{-0.2cm}
\end{table*}

\subsection{Implementation Details}
\subsubsection{LMM Training}
Following the mainstream choice of LMM-involved quality assessment methods \cite{wu2023qinstruct,wu2023qalign,coinstruct}, we select the mPLUG-Owl-2 \cite{mplugowl2} as the LMM model in this paper. 
The model comprises a CLIP-ViT-Large \cite{clip} visual encoder $\mathcal{E}_v$ with 304 million parameters, a visual abstractor $\mathcal{\hat E}_v$ with 82 million parameters, and the LLaMA2-7B \cite{llama2} LLM $\mathcal{L}$ on top of the visual modules, which integrates an additional multi-way module from mPLUG-Owl2, totaling 7.8 billion parameters. Input projections are initially squared through padding before being resized to $448\times448$. Let $\mathcal{E}_t$ represent the text embedding layer, with input projections denoted as \textit{<img1><img2>...} and the text prompt as $t$, the detailed formulation of the used LMM model can be expressed as follows:
\begin{equation}
    \begin{aligned}
    \mathcal{H}_{v}&= \mathcal{\hat E}_v(\mathcal{E}_v(<img1><img2>...)),\\
    \mathcal{H}_{t}&= \mathcal{E}_t(t),\\
    \mathcal{H} &= \mathcal{H}_{v}\oplus\mathcal{H}_{t},  \\
    \mathcal{O} &= \mathcal{L}(\mathcal{H}),
    \end{aligned}
\end{equation}
where $\mathcal{H}_{v}$ and $\mathcal{H}_{t}$ represent the abstracted tokens for the visual and text input respectively, $\mathcal{O}$ stands for the output.
For all PCQA databases, the batch size is maintained at 64. The learning rate is fixed at $2 \times 10^{-5}$, with the training process extending over 2 epochs for each variant. We utilize the common GPT~\cite{gpt2} loss mechanism, specifically the cross-entropy between the predicted logits and actual labels. The evaluation of performance metrics is conducted using the final weights obtained post-training. Four NVIDIA A100 80G GPUs are employed for the training phase, while a single RTX3090 24G GPU is used to measure inference latency.
In the inference stage, only the input texts preceding the \textit{[SCORE\_TOKEN]} are inputted into the LMM, leading to the final element of $\mathbf{O}$ representing the targeted probability map.

\subsubsection{Validation Strategy}

Following the methodologies in \cite{fan2022no,zhang2022mm,chai2024plain}, we utilize a k-fold cross-validation approach in our experiments to ensure a dependable performance evaluation of our proposed method. The SJTU-PCQA, WPC, and WPC2.0 databases consist of 9, 20, and 16 point cloud groups, respectively, leading us to adopt 9-fold, 5-fold, and 4-fold cross-validation for these databases to achieve an approximate 8:2 train-test split. The average of the performance metrics is considered the definitive result.
It is crucial to note that the training and testing sets are mutually exclusive to prevent content overlap. For FR-PCQA methods, which do not require training, we evaluate them using the same test sets and report the average performance.

% The RR-PCQA method includes PCMRR \cite{viola2020pcmrr}.
\subsection{Competitors}
17 state-of-the-art quality assessment methods are selected for comparison, which consists of 8 FR-PCQA and 9 NR-PCQA methods: 
\begin{itemize}
    \item The FR-PCQA methods include MSE-p2point (MSE-p2po) \cite{mekuria2016evaluation}, Hausdorff-p2point (HD-p2po) \cite{mekuria2016evaluation}, MSE-p2plane (MSE-p2pl) \cite{tian2017geometric}, Hausdorff-p2plane (HD-p2pl) \cite{tian2017geometric}, PSNR-yuv \cite{torlig2018novel}, PCQM \cite{meynet2020pcqm}, GraphSIM \cite{yang2020graphsim}, and PointSSIM \cite{alexiou2020pointssim}. 
    \item The NR-PCQA methods include BRISQUE \cite{mittal2012brisque}, NIQE \cite{mittal2012making}, IL-NIQE \cite{zhang2015feature}, IT-PCQA \cite{yang2022no}, ResSCNN \cite{liu2022point}, PQA-net \cite{liu2021pqa}, 3D-NSS \cite{zhang2021no}, GMS-3DQA \cite{zhang2023gms} and MM-PCQA \cite{zhang2022mm}. 
\end{itemize} 
Note that BRISQUE, NIQE, IL-NIQE are image-based quality assessment metrics and are validated on the same projections.
% To reconcile the scale discrepancies between predicted quality scores and quality labels, a five-parameter logistic function is employed for mapping, following the recommendation in \cite{antkowiak2000final}.

\subsection{Evaluation Criteria}

Four mainstream evaluation criteria in the quality assessment field are utilized to compare the correlation between the predicted scores and MOSs, which include Spearman Rank Correlation Coefficient (SRCC), Kendall’s Rank Correlation Coefficient (KRCC), Pearson Linear Correlation Coefficient (PLCC), Root Mean Squared Error (RMSE). 
An excellent quality assessment model should obtain values of SRCC, KRCC, PLCC close to 1 and RMSE to 0.

\subsection{Performance Discussion}
The performance comparison between the proposed LMM-PCQA and other PCQA competitors on the SJTU-PCQA, WPC, and WPC2.0 databases is illustrated in Table \ref{tab:experiment}, from which we can draw several conclusions: 1) The LMM-PCQA demonstrates superior performance across all three PCQA databases, even outperforming FR-PCQA methods. For instance, the proposed LMM-PCQA surpasses the second-best NR-PCQA method by about 0.027 (against GMS-3DQA), 0.041 (against MM-PCQA), and 0.034 (against GMS-3DQA) on the SJTU-PCQA, WPC, and WPC2.0 databases from the SRCC values. This highlights LMM's capability to effectively assimilate PCQA knowledge and apply it actively. The consistency in LMM-PCQA's performance across various databases underscores its potential to set new baselines in the PCQA field. 2) All PCQA competitors generally perform better on the SJTU-PCQA database but face significant performance declines on the WPC and WPC2.0 databases. In contrast, LMM-PCQA exhibits the smallest performance drops, with decreases of approximately 0.06 and 0.08 in SRCC values when transitioning from the SJTU-PCQA to the WPC and WPC2.0 databases, respectively. This performance stability underscores LMM-PCQA's robustness and superior ability to handle diverse content effectively, showcasing its adaptability and consistency in quality assessment across different point cloud databases.

% 1) The LMM-PCQA demonstrates superior performance across all three PCQA databases, even outperforming FR-PCQA methods. This highlights the capability of LMM to assimilate PCQA knowledge effectively and apply it actively. 2) 
% All PCQA competitors generally perform better on the SJTU-PCQA database but face significant performance declines on the WPC and WPC2.0 databases. In contrast, the proposed LMM-PCQA exhibits the smallest performance drops, with decreases of approximately 0.06 and 0.08 in SRCC values when transitioning from the SJTU-PCQA database to the WPC and WPC2.0 databases, respectively. This performance stability underscores LMM-PCQA's robustness and superior ability to handle diverse content effectively.

\begin{table}[tb]
   \caption{Contributions of LMM evaluation results and structural features, where `w/o LMM' indicates excluding the LMM evaluation results, `w/o Structural' indicates excluding the structural features. }
\renewcommand\tabcolsep{2pt}
    \centering
    \begin{tabular}{c|cc|cc|cc} 
    \toprule
    \multirow{2}{*}{Modal}  & \multicolumn{2}{c|}{SJTU-PCQA} & \multicolumn{2}{c|}{WPC} & \multicolumn{2}{c}{WPC2.0} \\ \cline{2-7}
            & SRCC$\uparrow$    & PLCC$\uparrow$   & SRCC$\uparrow$   & PLCC$\uparrow$ & SRCC$\uparrow$   & PLCC$\uparrow$ \\ \hline
        w/o LMM          &0.6650 &0.7274 & 0.3598 & 0.3523 & 0.3847 & 0.3951\\ 
        w/o Structural         & \bluebold{0.9081} & \bluebold{0.9158} & \bluebold{0.8488} & \bluebold{0.8271} & \bluebold{0.8258} & \bluebold{0.8381}\\
        LMM + Structural      &\redbold{0.9376} &\redbold{0.9404} & \redbold{0.8825} & \redbold{0.8739} &\redbold{0.8614} & \redbold{0.8634} \\     
    \bottomrule
    \end{tabular}
    \label{tab:modality}
    \vspace{-0.2cm}
\end{table}

\begin{figure}[t]
    \centering
    \includegraphics[width=.8\linewidth]{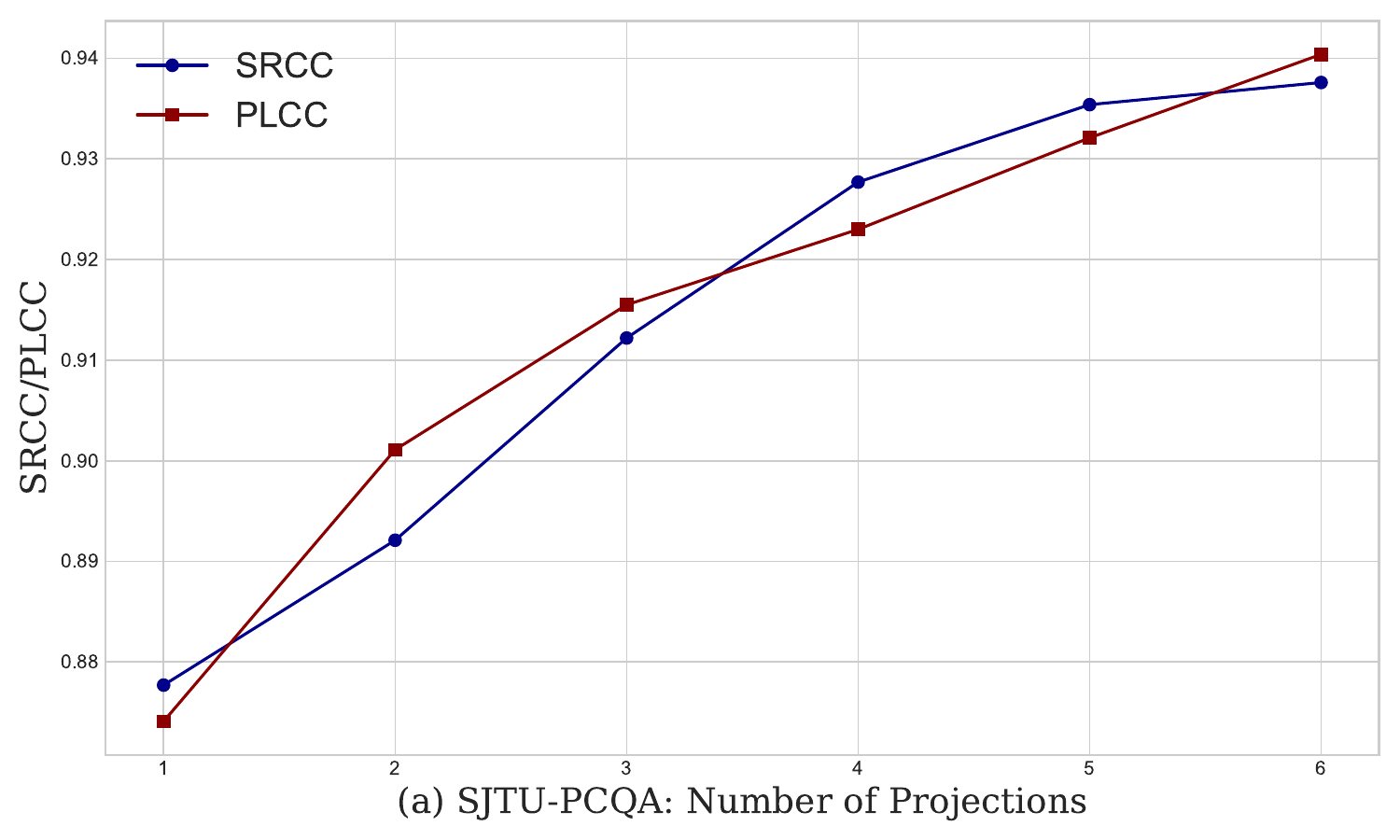}
    \includegraphics[width=.8\linewidth]{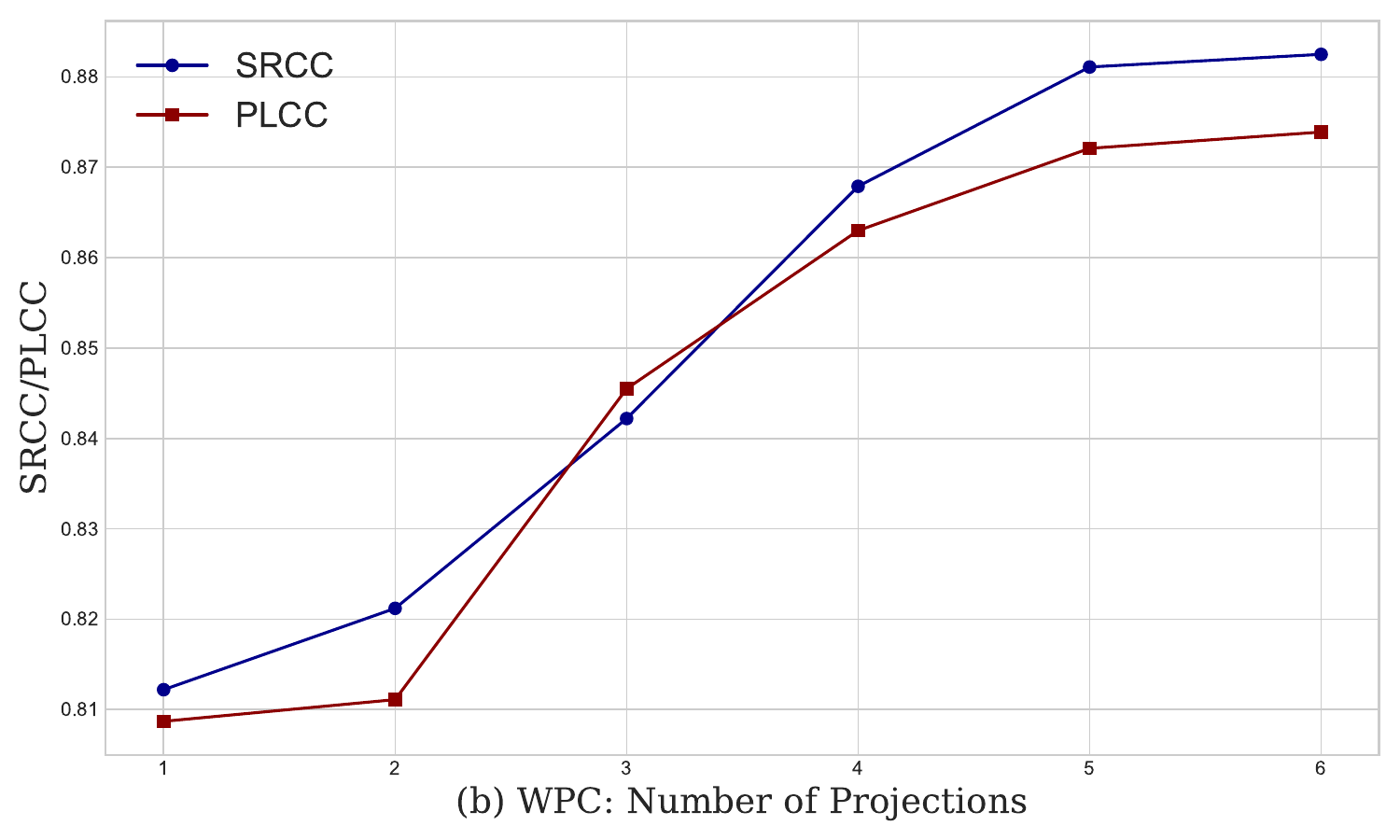}
    \includegraphics[width=.8\linewidth]{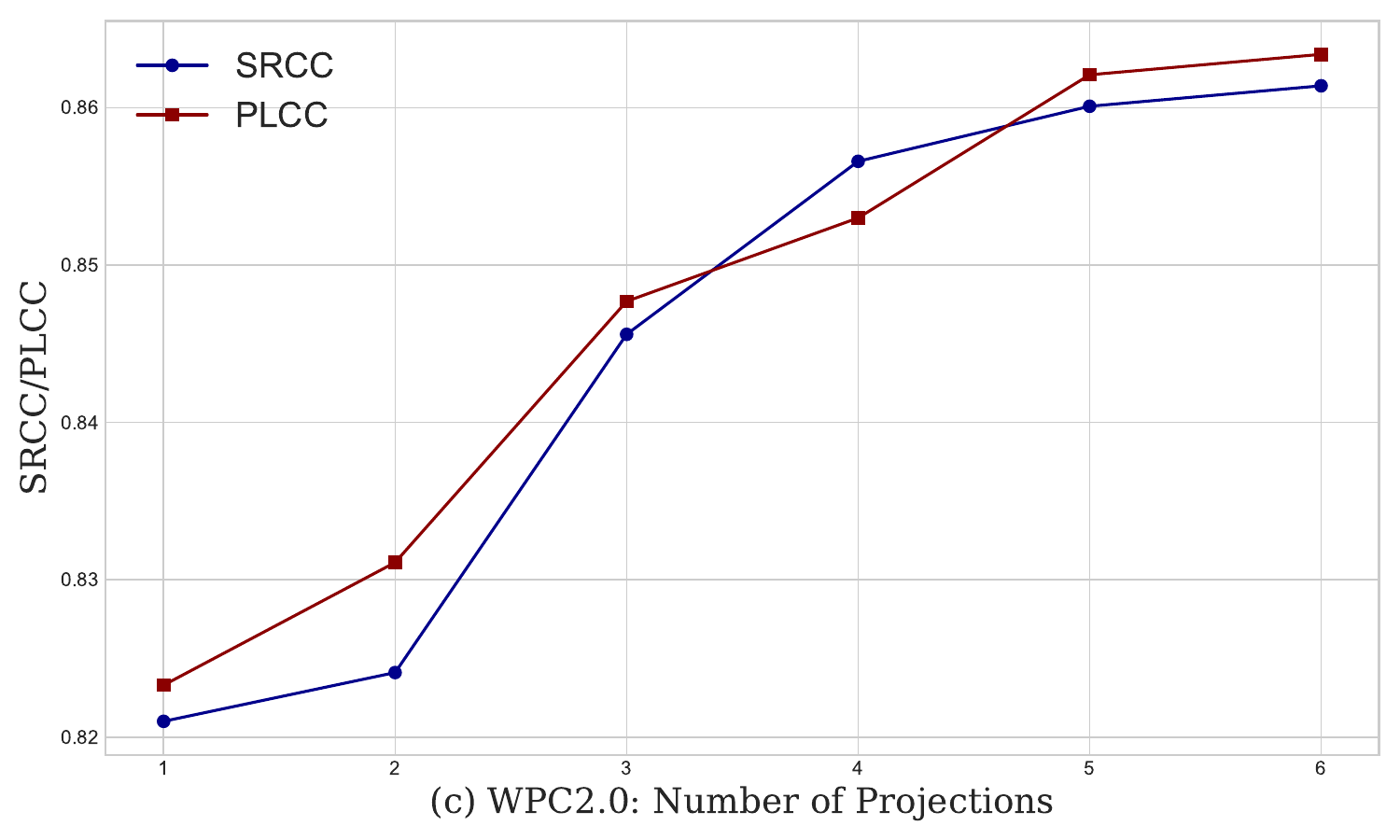}
    \caption{SRCC/PLCC performance tendency according to the number of used projections on the SJTU-PCQA, WPC, and WPC2.0 databases.}
    \label{fig:projection}
\end{figure}

\subsection{Ablation Study}
\subsubsection{Contributions of LMM evaluation results and structural features}
To fully investigate the contributions and validate the rationality behind the proposed dual streams of features, we decide to undertake an ablation study in this section. The results, as detailed in Table \ref{tab:modality}, clearly demonstrate that the integration of both feature streams leads to superior performance compared to employing a single feature stream. Upon a detailed examination, it is apparent that LMM evaluation results markedly surpass the structural features in terms of performance. This disparity primarily stems from the fact that some distorted point clouds within the SJTU-PCQA, WPC, and WPC2.0 databases suffer only from color distortions. Structural features, derived from geometric information, inherently lack the capability to recognize these color distortions, resulting in their relatively lower performance in total. However, incorporating structural features enhances the comprehensive understanding of point cloud quality, contributing to the refinement and elevation of the assessment precision in LMM evaluation results. This combined method highlights how important it is to look at features from different angles to fully understand and capture the subtle details of point cloud quality.

% This synergistic approach underscores the importance of a multi-faceted feature analysis in capturing the full spectrum of quality nuances within point cloud data.

\subsubsection{Influence of the number of projections}
The proposed LMM-PCQA utilizes 6 projections as default. In this section, we further change the number of used projections to test the corresponding performance influence. Specifically, we randomly select 1-6 projections from the cube-like 6 projections setting as the input projections of LMM-PCQA. The performance tendency is illustrated in Fig. \ref{fig:projection}, from which we can draw several interesting conclusions: 1) As the number of projections increases, the performance of LMM-PCQA also improves correspondingly, indicating that increasing the number of projections can encompass more effective quality information, thereby enhancing the final performance. 2) Specifically, when the number of projections increases from 2 to 5, the improvement is significantly more pronounced. This suggests that at this stage, the quality information is not yet redundant, and the benefit of increasing the number of projections is relatively large. However, when the number of projections increases from 5 to 6, the performance improvement is relatively low, indicating that the quality information has become somewhat saturated, and further increases in the number of projections yield diminishing returns.

\subsubsection{Effect of the multi-scale structural features}
To quantify the contributions of the multi-scale mechanism, we validate the performance of structural features with different scale parameters under two settings: with LMM evaluation results and without LMM evaluation results. The experimental performance is listed in Table \ref{tab:multiscale}. From the table, we can find that the multi-scale structural features with k=10,20 perform better than the single-scale features whether the LMM evaluation results are involved or not, which confirms the effectiveness of the proposed multi-scale mechanism. This can be attributed to that humans tend to perceive the visual quality of point clouds from a multi-scale perspective.

\begin{table}[tb]
\caption{Performance of the multi-scale structural features, where k is the scale parameter of the KNN algorithm. }
\renewcommand\tabcolsep{4pt}
    \centering
    \begin{tabular}{c|cc|cc|cc}
    \toprule
    \multirow{2}{*}{Model}  & \multicolumn{2}{c|}{SJTU-PCQA} & \multicolumn{2}{c|}{WPC} & \multicolumn{2}{c}{WPC2.0}\\ \cline{2-7}
            & SRCC$\uparrow$    & PLCC$\uparrow$   & SRCC$\uparrow$   & PLCC$\uparrow$ & SRCC$\uparrow$   & PLCC$\uparrow$\\ \hline
        \multicolumn{7}{l}{w/o LMM} \\ \hdashline
        k=10         &\bluebold{0.6090} & \bluebold{0.6584} & \bluebold{0.3261} & \bluebold{0.3482} & \bluebold{0.3366} & 0.2777\\
        k=20         &0.5920 &0.6311 & 0.1795 & 0.2720 & 0.3224 & \bluebold{0.3129}\\ 
        k={10,20}  &\redbold{0.6650} &\redbold{0.7274} & \redbold{0.3598} & \redbold{0.3523} & \redbold{0.3847} & \redbold{0.3951}\\ \hdashline
        \multicolumn{7}{l}{with LMM} \\ \hdashline     
        k=10         &0.9140 &0.9176 & \bluebold{0.8564} & \bluebold{0.8554} & 0.8432 & 0.8466\\
        k=20     &\bluebold{0.9199} & \bluebold{0.9179} & 0.8466 & 0.8488 & \bluebold{0.8578} & \bluebold{0.8562}\\ 
        k={10,20} &\redbold{0.9376} &\redbold{0.9404} & \redbold{0.8825} & \redbold{0.8739} &\redbold{0.8614} & \redbold{0.8634} \\
    \bottomrule
    \end{tabular}
    \vspace{-0.2cm}
    \label{tab:multiscale}
\end{table}

\begin{table}[t]
    \centering
    \caption{The cross-database evaluation performance, `WPC$\rightarrow$SJTU-PCQA' signifies that the model is trained using the WPC database and tested according to the standard testing protocol of the SJTU-PCQA database. We eliminate those point cloud groups from the WPC database that have reference counterparts in the WPC2.0 testing sets, thereby preventing content duplication.}
    \renewcommand\tabcolsep{5pt}
    \begin{tabular}{c|cc|cc}
    \toprule
    \multirow{2}{*}{Model}  & \multicolumn{2}{c|}{WPC$\rightarrow$SJTU-PCQA} & \multicolumn{2}{c}{WPC$\rightarrow$WPC2.0} \\ \cline{2-5}
            & SRCC$\uparrow$    & PLCC$\uparrow$   & SRCC$\uparrow$   & PLCC$\uparrow$ \\ \hline
        PQA-net       & {0.5411} & {0.6102} & {0.6006} & {0.6377}\\
        3D-NSS        & 0.1817 & 0.2344 & 0.4933 & 0.5613\\ 
        GMS-3DQA      & 0.7421 & 0.7611 & 0.7822 & 0.7714 \\
        MM-PCQA       & \bluebold{0.7991}   & \bluebold{0.7902}    & \bluebold{0.7917}    & \bluebold{0.7935}\\    
        LMM-PCQA(Ours) & \redbold{0.8246}   & \redbold{0.7999}    &\redbold{0.8385}    & \redbold{0.8387}\\ 
    \bottomrule
    \end{tabular}
    
    \label{tab:crossdatabase}
    \vspace{-0.2cm}
\end{table}

\begin{table*}[!t]
\setlength\tabcolsep{5.5pt}
\renewcommand\arraystretch{0.95}
\caption{Distortion-specific performance results on the SJTU-PCQA database, where OT represents octree-based compression, CN represents color noise, DS represents down-sampling, DS+CN represents down-sampling and color noise, DS+GGN represents down-sampling and geometry Gaussian noise, GGN represents geometry Gaussian noise, and CN+GGN represents color noise and geometry Gaussian noise respectively. }
\centering
\begin{tabular}{l|c|c|c|c|c|c|c|c|c|c|c|c|c|c}
\toprule
Distortion    & \multicolumn{2}{c|}{OT}        & \multicolumn{2}{c|}{CN}        & \multicolumn{2}{c|}{DS}        & \multicolumn{2}{c|}{DS+CN}     & \multicolumn{2}{c|}{DS+GGN}    & \multicolumn{2}{c|}{GGN}       & \multicolumn{2}{c}{CN+GGN}     \\ \hline
Method        & SR$\uparrow$          & PL$\uparrow$          & SR$\uparrow$         & PL$\uparrow$          & SR$\uparrow$          & PL$\uparrow$          & SR$\uparrow$          & PL$\uparrow$          & SR$\uparrow$         & PL$\uparrow$         & SR$\uparrow$         & PL$\uparrow$          & SR$\uparrow$          & PL$\uparrow$           \\ \hline
MSE-p2po      & 0.71          & 0.76          & nan           & nan           & 0.93          & 0.95          & \bluebold{0.96}          & 0.87          & {0.96} & 0.89          & \redbold{0.98} & 0.89          & \redbold{0.99} & 0.90       \\
HD-p2po       & 0.64          & 0.69          & nan           & nan           & 0.82          & 0.88          & 0.80          & 0.75          & 0.94          & 0.91          & \redbold{0.98} & 0.91          & \redbold{0.99} & 0.91       \\
MSE-p2pl      & 0.55          & 0.62          & nan           & nan           & 0.87          & 0.92          & 0.85          & 0.81          & {0.96} & 0.75          & \bluebold{0.97}          & 0.85          & \bluebold{0.98}          & 0.86        \\
HD-p2pl       & 0.54          & 0.58          & nan           & nan           & 0.82          & 0.87          & 0.81          & 0.79          & 0.94          & 0.77          & 0.95          & 0.88          & 0.97          & 0.85        \\
PSNR-yuv      & 0.59          & 0.54          & 0.86          & 0.87          & 0.91          & 0.91          & \bluebold{0.96}          & 0.91          & \bluebold{0.97} & 0.94          & \redbold{0.98} & 0.95          & \redbold{0.99} & 0.96         \\
PCQM          & 0.80          & 0.84          & 0.86          & 0.85          & 0.93          & 0.96          & \redbold{0.97} & \bluebold{0.94}          & {0.96} & 0.90          & \redbold{0.98} & 0.93   & \redbold{0.99} & 0.93       \\
GraphSIM      & {0.82}          & 0.81          & 0.82          & {0.90}          & \bluebold{0.96} & \bluebold{0.97} & 0.91          & \redbold{0.95} & 0.95          & 0.95          & 0.96          & \redbold{0.97} & 0.97          & \redbold{0.98} \\
PointSSIM     & 0.80          & \bluebold{0.88}          & {0.87}          & 0.87          & 0.93 & 0.93 & \redbold{0.97}          & 0.93          & {0.96} & \bluebold{0.97}          & \redbold{0.98} & \redbold{0.97} & \redbold{0.99} & 0.96     \\
PQA-net       & 0.81          & 0.82          & 0.84          & 0.83          & 0.91          & 0.92          & 0.93          & 0.91          & 0.89          & 0.89          & 0.95          & \bluebold{0.96}          & 0.97          & 0.96      \\
3D-NSS        & 0.60          & 0.67          & 0.85          & 0.79          & 0.80          & 0.84          & 0.94          & 0.93          & 0.90          & 0.90          & 0.96          & 0.93          & \bluebold{0.98}          & 0.94    \\
GMS-3DQA        & 0.83          & 0.84          & 0.91          & 0.92          & 0.95          & 0.95          & 0.95          & 0.93          & 0.96          & \bluebold{0.97}          & \bluebold{0.97}          & 0.93          & \bluebold{0.98}          & 0.95    \\
MM-PCQA        & \bluebold{0.84}          & 0.83          & \bluebold{0.92}          & \bluebold{0.93}          & 0.94          & 0.96          & 0.94          & \redbold{0.95}          & 0.94          & 0.93          & 0.95          & 0.94          & \bluebold{0.98}          & 0.96    \\
LMM-PCQA & \redbold{0.89} & \redbold{0.90} & \redbold{0.97} & \redbold{0.97} & \redbold{0.97}          & \redbold{0.98}          & 0.95 & \bluebold{0.94} & \redbold{0.98} & \redbold{0.98} & 0.96          & 0.93          & \bluebold{0.98}         & \bluebold{0.97}  \\ \bottomrule
\end{tabular}
\label{tab:sjtu}
\vspace{-0.3cm}
\end{table*}

\begin{table}[ht]
\centering
\setlength\tabcolsep{3.5pt}
\renewcommand\arraystretch{0.95}
\caption{Distortion-specific performance results on the WPC database, where DS represents down-sampling, GN represents geometry and color Gaussian noise, G-PCC represents geometry-based compression, and V-PCC represents video-based compression. }
\begin{tabular}{l|c|c|c|c|c|c|c|c}
\toprule
Distortion    & \multicolumn{2}{c|}{DS}        & \multicolumn{2}{c|}{GN}        & \multicolumn{2}{c|}{G-PCC}     & \multicolumn{2}{c}{V-PCC} \\ \hline
Method        & SR$\uparrow$          & PL$\uparrow$          & SR$\uparrow$          & PL$\uparrow$          & SR$\uparrow$          & PL$\uparrow$          & SR$\uparrow$          & PL$\uparrow$        \\ \hline
MSE-p2po      & 0.46          & 0.46          & 0.67          & 0.63          & 0.84          & 0.72          & 0.41          & 0.44             \\
HD-p2po       & 0.39          & 0.35          & 0.69          & 0.70          & 0.80          & 0.70          & 0.36          & 0.36           \\
MSE-p2pl      & 0.41          & 0.40          & 0.61          & 0.47          & 0.61          & 0.48          & 0.42          & 0.40           \\
HD-p2pl       & 0.40          & 0.41          & 0.62          & 0.51          & 0.57          & 0.59          & 0.41          & 0.41             \\
PSNR-yuv      & 0.23          & 0.28          & 0.79          & {0.88}          & 0.47          & 0.43          & 0.44          & 0.48           \\
PCQM          & {0.66}          & {0.64}          & \bluebold{0.89}          & \bluebold{0.89}          & 0.86          & 0.77          & {0.83}          & {0.78}         \\
GraphSIM      & 0.56          & 0.57          & 0.79          & 0.81          & 0.75          & 0.74          & 0.71          & 0.70          \\
PointSSIM     & 0.35          & 0.34          & 0.83          & 0.87          & \bluebold{0.91} & \bluebold{0.95} & 0.51          & 0.41          \\
PQA-net       & 0.61          & 0.63          & 0.77          & 0.78          & {0.87}          & 0.88          & 0.76          & 0.77      \\
3D-NSS        & 0.55          & 0.51          & 0.81          & 0.83          & 0.86          & 0.87          & 0.49          & 0.47            \\
GMS-3DQA        & 0.72          & 0.73          & 0.87          & {0.88}          & 0.89          & 0.89          & 0.86          & \bluebold{0.88}            \\
MM-PCQA        & \bluebold{0.75}          & \bluebold{0.74}          & {0.88}          & 0.87          & {0.88}          & 0.89          & \bluebold{0.89}          & 0.86            \\
LMM-PCQA & \redbold{0.76} & \redbold{0.79} & \redbold{0.91} & \redbold{0.91} & \redbold{0.96}          & \redbold{0.96}          & \redbold{0.90} & \redbold{0.89}  \\ \bottomrule
\end{tabular}
\label{tab:wpc}
\vspace{-0.3cm}
\end{table}

\subsection{Cross-database Validation}
To the generalization ability of the proposed LMM-PCQA, we conduct the cross-database validation in this section. Considering that the SJTU-PCQA, WPC, and WPC2.0 databases contain 378, 740, and 400 distorted point clouds respectively, we pre-train LMM-PCQA on the WPC database (largest in scale) and validate the performance on the SJTU-PCQA and WPC2.0 databases (smaller in scale). The competitive NR-PCQA methods (PQA-net, 3d-NSS, GMS-3DQA, and MM-PCQA) are included for comparison. The experimental performance is shown in Table \ref{tab:crossdatabase}, from which we can make several observations: 1) The proposed LMM-PCQA achieves the best cross-database validation performance against all competitors, which confirms the strong generalization ability of LMM-PCQA. 2) Most methods obtain higher WPC$\rightarrow$WPC2.0 performance than WPC$\rightarrow$SJTU-PCQA performance. This might be because the WPC2.0 database contains only compression distortions, which undergo part of a similar distortion generation process of the WPC database. Therefore, the quality representation learned from the WPC database is more effective on the WPC2.0 database.

\subsection{Distortion-specific Evaluation}
To verify the effectiveness of the proposed LMM-PCQA on different kinds of distortions, we carry out the distortion-specific evaluation experiment in this section. The performance comparison is shown in Table \ref{tab:sjtu} (The `nan' values for MSE-p2po, HD-p2po, MSE-p2pl, and HD-p2pl arise because these methods only analyze the geometric differences between the reference and distorted point clouds, thereby failing to account for color distortions.) and Table \ref{tab:wpc}, from which we can obtain several findings: 1) The proposed LMM-PCQA achieves the best performance on 4 of 7 distortion types and all distortion types on the SJTU-PCQA database and the WPC database respectively, which suggests that LMM-PCQA is effective at dealing with various kinds of distortions. 2) Although LMM-PCQA does not achieve the top performance on DS+CN, GGN, and CN+GGN distortions within the SJTU-PCQA database, the performance gap to the best is minimal, with a difference of no more than 0.02 in terms of SRCC values. This suggests that LMM-PCQA remains in the top tier. 3) Upon examining Table \ref{tab:wpc} more closely, it is evident that all PCQA methods exhibit substantial declines in performance with the `DS' distortion. The underlying reason for this trend is the simplicity of the point cloud reference models used in the WPC database, leading to a reduced sensitivity to downsampling distortion.
% \subsection{Visualization Results}

% \begin{figure*}
%     \centering
%     \fbox{\phantom{\rule{16cm}{5cm}}} 
%     \caption{The visualization results.}
%     \label{fig:visulization}
% \end{figure*}
% \vspace{-0.2cm}

\section{Conclusion}
In conclusion, this paper pioneers the integration of LMMs with PCQA, unveiling the untapped potential of LMMs in this domain. Our research successfully demonstrates the feasibility of adapting LMMs for PCQA through text supervision, enhancing their capability to evaluate 3D visual quality from 2D projections. 
We also propose to capture multi-scale structural features to offer a more holistic view of the point cloud quality. By combining LMM evaluation results and structural features, our approach significantly improves the accuracy of PCQA. We carry out thorough experiments to demonstrate the effectiveness of the proposed LMM-PCQA, as well as its robustness and ability to generalize across various distortion types and diverse point cloud content. The encouraging results not only validate our methodology but also lay the groundwork for future research. We hope our work will serve as a stepping stone for further exploration into the synergy between LMMs and PCQA, driving forward innovations in 3D visual quality analysis.
%%
%% The acknowledgments section is defined using the "acks" environment
%% (and NOT an unnumbered section). This ensures the proper
%% identification of the section in the article metadata, and the
%% consistent spelling of the heading.
\begin{acks}
This work was supported in part by the National Natural Science Foundation of China (623B2073, 62101326, 62225112, 62301316), the Foundation of Key Laboratory of Media Audio \& Video (Communication University of China), Ministry of Education, China (JGKFKT2302), the China Postdoctoral Science Foundation under Grants 2023TQ0212 and 2023M742298, and the Postdoctoral Fellowship Program of CPSF under Grant GZC20231618. 
\end{acks}

%%
%% The next two lines define the bibliography style to be used, and
%% the bibliography file.
\bibliographystyle{ACM-Reference-Format}
\bibliography{sample-base}

%%% -*-BibTeX-*-
%%% Do NOT edit. File created by BibTeX with style
%%% ACM-Reference-Format-Journals [18-Jan-2012].

\begin{thebibliography}{65}

%%% ====================================================================
%%% NOTE TO THE USER: you can override these defaults by providing
%%% customized versions of any of these macros before the \bibliography
%%% command.  Each of them MUST provide its own final punctuation,
%%% except for \shownote{}, \showDOI{}, and \showURL{}.  The latter two
%%% do not use final punctuation, in order to avoid confusing it with
%%% the Web address.
%%%
%%% To suppress output of a particular field, define its macro to expand
%%% to an empty string, or better, \unskip, like this:
%%%
%%% \newcommand{\showDOI}[1]{\unskip}   % LaTeX syntax
%%%
%%% \def \showDOI #1{\unskip}           % plain TeX syntax
%%%
%%% ====================================================================

\ifx \showCODEN    \undefined \def \showCODEN     #1{\unskip}     \fi
\ifx \showDOI      \undefined \def \showDOI       #1{#1}\fi
\ifx \showISBNx    \undefined \def \showISBNx     #1{\unskip}     \fi
\ifx \showISBNxiii \undefined \def \showISBNxiii  #1{\unskip}     \fi
\ifx \showISSN     \undefined \def \showISSN      #1{\unskip}     \fi
\ifx \showLCCN     \undefined \def \showLCCN      #1{\unskip}     \fi
\ifx \shownote     \undefined \def \shownote      #1{#1}          \fi
\ifx \showarticletitle \undefined \def \showarticletitle #1{#1}   \fi
\ifx \showURL      \undefined \def \showURL       {\relax}        \fi
% The following commands are used for tagged output and should be
% invisible to TeX
\providecommand\bibfield[2]{#2}
\providecommand\bibinfo[2]{#2}
\providecommand\natexlab[1]{#1}
\providecommand\showeprint[2][]{arXiv:#2}

\bibitem[itu(2000)]%
        {itu}
 \bibinfo{year}{2000}\natexlab{}.
\newblock \bibinfo{title}{Recommendation 500-10: Methodology for the subjective assessment of the quality of television pictures}.
\newblock \bibinfo{howpublished}{ITU-R Rec. BT.500}.
\newblock


\bibitem[Alexiou et~al\mbox{.}(2019)]%
        {alexiou2019exploiting}
\bibfield{author}{\bibinfo{person}{Evangelos Alexiou} {et~al\mbox{.}}} \bibinfo{year}{2019}\natexlab{}.
\newblock \showarticletitle{Exploiting user interactivity in quality assessment of point cloud imaging}. In \bibinfo{booktitle}{\emph{QoMEX}}. IEEE.
\newblock


\bibitem[Alexiou and Ebrahimi(2020)]%
        {alexiou2020pointssim}
\bibfield{author}{\bibinfo{person}{Evangelos Alexiou} {and} \bibinfo{person}{Touradj Ebrahimi}.} \bibinfo{year}{2020}\natexlab{}.
\newblock \showarticletitle{Towards a point cloud structural similarity metric}. In \bibinfo{booktitle}{\emph{International Conference on Multimedia and Expo Workshop}}. \bibinfo{pages}{1--6}.
\newblock


\bibitem[Alexiou et~al\mbox{.}(2021)]%
        {alexiou2021pointpca}
\bibfield{author}{\bibinfo{person}{Evangelos Alexiou}, \bibinfo{person}{Xuemei Zhou}, \bibinfo{person}{Irene Viola}, {and} \bibinfo{person}{Pablo Cesar}.} \bibinfo{year}{2021}\natexlab{}.
\newblock \showarticletitle{PointPCA: Point cloud objective quality assessment using PCA-based descriptors}.
\newblock \bibinfo{journal}{\emph{arXiv preprint arXiv:2111.12663}} (\bibinfo{year}{2021}).
\newblock


\bibitem[Chai et~al\mbox{.}(2024)]%
        {chai2024plain}
\bibfield{author}{\bibinfo{person}{Xiongli Chai}, \bibinfo{person}{Feng Shao}, \bibinfo{person}{Baoyang Mu}, \bibinfo{person}{Hangwei Chen}, \bibinfo{person}{Qiuping Jiang}, {and} \bibinfo{person}{Yo-Sung Ho}.} \bibinfo{year}{2024}\natexlab{}.
\newblock \showarticletitle{Plain-PCQA: No-Reference Point Cloud Quality Assessment by Analysis of Plain Visual and Geometrical Components}.
\newblock \bibinfo{journal}{\emph{IEEE Transactions on Circuits and Systems for Video Technology}} (\bibinfo{year}{2024}).
\newblock


\bibitem[Chen et~al\mbox{.}(2020)]%
        {chen2020object}
\bibfield{author}{\bibinfo{person}{Qi Chen}, \bibinfo{person}{Lin Sun}, \bibinfo{person}{Zhixin Wang}, \bibinfo{person}{Kui Jia}, {and} \bibinfo{person}{Alan Yuille}.} \bibinfo{year}{2020}\natexlab{}.
\newblock \showarticletitle{Object as hotspots: An anchor-free 3d object detection approach via firing of hotspots}. In \bibinfo{booktitle}{\emph{European Conference on Computer Vision}}. \bibinfo{pages}{68--84}.
\newblock


\bibitem[Chen(2023)]%
        {chen2023x}
\bibfield{author}{\bibinfo{person}{Yixiong Chen}.} \bibinfo{year}{2023}\natexlab{}.
\newblock \showarticletitle{X-IQE: eXplainable Image Quality Evaluation for Text-to-Image Generation with Visual Large Language Models}.
\newblock \bibinfo{journal}{\emph{arXiv preprint arXiv:2305.10843}} (\bibinfo{year}{2023}).
\newblock


\bibitem[Cheng et~al\mbox{.}(2021)]%
        {cheng2021sspc}
\bibfield{author}{\bibinfo{person}{Mingmei Cheng}, \bibinfo{person}{Le Hui}, \bibinfo{person}{Jin Xie}, {and} \bibinfo{person}{Jian Yang}.} \bibinfo{year}{2021}\natexlab{}.
\newblock \showarticletitle{SSPC-Net: Semi-supervised semantic 3D point cloud segmentation network}. In \bibinfo{booktitle}{\emph{AAAI}}.
\newblock


\bibitem[Chetouani et~al\mbox{.}(2021)]%
        {chetouani2021deep}
\bibfield{author}{\bibinfo{person}{Aladine Chetouani}, \bibinfo{person}{Maurice Quach}, \bibinfo{person}{Giuseppe Valenzise}, {and} \bibinfo{person}{Fr{\'e}d{\'e}ric Dufaux}.} \bibinfo{year}{2021}\natexlab{}.
\newblock \showarticletitle{Deep learning-based quality assessment of 3d point clouds without reference}. In \bibinfo{booktitle}{\emph{International Conference on Multimedia and Expo Workshop}}. \bibinfo{pages}{1--6}.
\newblock


\bibitem[Cui et~al\mbox{.}(2021)]%
        {cui2021deep}
\bibfield{author}{\bibinfo{person}{Yaodong Cui}, \bibinfo{person}{Ren Chen}, \bibinfo{person}{Wenbo Chu}, \bibinfo{person}{Long Chen}, \bibinfo{person}{Daxin Tian}, \bibinfo{person}{Ying Li}, {and} \bibinfo{person}{Dongpu Cao}.} \bibinfo{year}{2021}\natexlab{}.
\newblock \showarticletitle{Deep learning for image and point cloud fusion in autonomous driving: A review}.
\newblock \bibinfo{journal}{\emph{IEEE Transactions on Intelligent Transportation Systems}} \bibinfo{volume}{23}, \bibinfo{number}{2} (\bibinfo{year}{2021}), \bibinfo{pages}{722--739}.
\newblock


\bibitem[Fan et~al\mbox{.}(2022a)]%
        {fan2022d}
\bibfield{author}{\bibinfo{person}{Tingyu Fan}, \bibinfo{person}{Linyao Gao}, \bibinfo{person}{Yiling Xu}, \bibinfo{person}{Zhu Li}, {and} \bibinfo{person}{Dong Wang}.} \bibinfo{year}{2022}\natexlab{a}.
\newblock \showarticletitle{D-DPCC: Deep Dynamic Point Cloud Compression via 3D Motion Prediction}.
\newblock \bibinfo{journal}{\emph{International Joint Conference on Artificial Intelligence}} (\bibinfo{year}{2022}).
\newblock


\bibitem[Fan et~al\mbox{.}(2022b)]%
        {fan2022no}
\bibfield{author}{\bibinfo{person}{Yu Fan}, \bibinfo{person}{Zicheng Zhang}, \bibinfo{person}{Wei Sun}, \bibinfo{person}{Xiongkuo Min}, \bibinfo{person}{Ning Liu}, \bibinfo{person}{Quan Zhou}, \bibinfo{person}{Jun He}, \bibinfo{person}{Qiyuan Wang}, {and} \bibinfo{person}{Guangtao Zhai}.} \bibinfo{year}{2022}\natexlab{b}.
\newblock \showarticletitle{A no-reference quality assessment metric for point cloud based on captured video sequences}. In \bibinfo{booktitle}{\emph{IEEE MMSP}}. IEEE, \bibinfo{pages}{1--5}.
\newblock


\bibitem[Grilli et~al\mbox{.}(2017)]%
        {grilli2017review}
\bibfield{author}{\bibinfo{person}{Eleonora Grilli}, \bibinfo{person}{Fabio Menna}, {and} \bibinfo{person}{Fabio Remondino}.} \bibinfo{year}{2017}\natexlab{}.
\newblock \showarticletitle{A review of point clouds segmentation and classification algorithms}.
\newblock \bibinfo{journal}{\emph{The International Archives of Photogrammetry, Remote Sensing and Spatial Information Sciences}}  \bibinfo{volume}{42} (\bibinfo{year}{2017}), \bibinfo{pages}{339}.
\newblock


\bibitem[Gu et~al\mbox{.}(2017)]%
        {gu2017learning}
\bibfield{author}{\bibinfo{person}{Ke Gu}, \bibinfo{person}{Dacheng Tao}, \bibinfo{person}{Jun-Fei Qiao}, {and} \bibinfo{person}{Weisi Lin}.} \bibinfo{year}{2017}\natexlab{}.
\newblock \showarticletitle{Learning a no-reference quality assessment model of enhanced images with big data}.
\newblock \bibinfo{journal}{\emph{IEEE Transactions on Neural Networks and Learning Systems}} \bibinfo{volume}{29}, \bibinfo{number}{4} (\bibinfo{year}{2017}), \bibinfo{pages}{1301--1313}.
\newblock


\bibitem[Gu et~al\mbox{.}(2019)]%
        {gu20193d}
\bibfield{author}{\bibinfo{person}{Shuai Gu}, \bibinfo{person}{Junhui Hou}, \bibinfo{person}{Huanqiang Zeng}, \bibinfo{person}{Hui Yuan}, {and} \bibinfo{person}{Kai-Kuang Ma}.} \bibinfo{year}{2019}\natexlab{}.
\newblock \showarticletitle{3D point cloud attribute compression using geometry-guided sparse representation}.
\newblock \bibinfo{journal}{\emph{IEEE Transactions on Image Processing}}  \bibinfo{volume}{29} (\bibinfo{year}{2019}), \bibinfo{pages}{796--808}.
\newblock


\bibitem[Huang et~al\mbox{.}(2024)]%
        {visualcritic}
\bibfield{author}{\bibinfo{person}{Zhipeng Huang}, \bibinfo{person}{Zhizheng Zhang}, \bibinfo{person}{Yiting Lu}, \bibinfo{person}{Zheng-Jun Zha}, \bibinfo{person}{Zhibo Chen}, {and} \bibinfo{person}{Baining Guo}.} \bibinfo{year}{2024}\natexlab{}.
\newblock \bibinfo{title}{VisualCritic: Making LMMs Perceive Visual Quality Like Humans}.
\newblock
\newblock
\showeprint[arxiv]{2403.12806}~[cs.CV]


\bibitem[Ku et~al\mbox{.}(2018)]%
        {ku2018joint}
\bibfield{author}{\bibinfo{person}{Jason Ku}, \bibinfo{person}{Melissa Mozifian}, \bibinfo{person}{Jungwook Lee}, \bibinfo{person}{Ali Harakeh}, {and} \bibinfo{person}{Steven~L Waslander}.} \bibinfo{year}{2018}\natexlab{}.
\newblock \showarticletitle{Joint 3d proposal generation and object detection from view aggregation}. In \bibinfo{booktitle}{\emph{IEEE/RSJ International Conference on Intelligent Robots and Systems}}. \bibinfo{pages}{1--8}.
\newblock


\bibitem[Liu et~al\mbox{.}(2023)]%
        {llava}
\bibfield{author}{\bibinfo{person}{Haotian Liu}, \bibinfo{person}{Chunyuan Li}, \bibinfo{person}{Qingyang Wu}, {and} \bibinfo{person}{Yong~Jae Lee}.} \bibinfo{year}{2023}\natexlab{}.
\newblock \bibinfo{title}{Visual Instruction Tuning}.
\newblock
\newblock


\bibitem[Liu et~al\mbox{.}(2022b)]%
        {liu2022perceptual}
\bibfield{author}{\bibinfo{person}{Qi Liu}, \bibinfo{person}{Honglei Su}, \bibinfo{person}{Zhengfang Duanmu}, \bibinfo{person}{Wentao Liu}, {and} \bibinfo{person}{Zhou Wang}.} \bibinfo{year}{2022}\natexlab{b}.
\newblock \showarticletitle{Perceptual Quality Assessment of Colored 3D Point Clouds}.
\newblock \bibinfo{journal}{\emph{IEEE Transactions on Visualization and Computer Graphics}} (\bibinfo{year}{2022}).
\newblock


\bibitem[Liu et~al\mbox{.}(2021a)]%
        {liu2021reduced}
\bibfield{author}{\bibinfo{person}{Qi Liu}, \bibinfo{person}{Hui Yuan}, \bibinfo{person}{Raouf Hamzaoui}, \bibinfo{person}{Honglei Su}, \bibinfo{person}{Junhui Hou}, {and} \bibinfo{person}{Huan Yang}.} \bibinfo{year}{2021}\natexlab{a}.
\newblock \showarticletitle{Reduced reference perceptual quality model with application to rate control for video-based point cloud compression}.
\newblock \bibinfo{journal}{\emph{IEEE Transactions on Image Processing}}  \bibinfo{volume}{30} (\bibinfo{year}{2021}), \bibinfo{pages}{6623--6636}.
\newblock


\bibitem[Liu et~al\mbox{.}(2020)]%
        {liu2020model}
\bibfield{author}{\bibinfo{person}{Qi Liu}, \bibinfo{person}{Hui Yuan}, \bibinfo{person}{Junhui Hou}, \bibinfo{person}{Raouf Hamzaoui}, {and} \bibinfo{person}{Honglei Su}.} \bibinfo{year}{2020}\natexlab{}.
\newblock \showarticletitle{Model-based joint bit allocation between geometry and color for video-based 3D point cloud compression}.
\newblock \bibinfo{journal}{\emph{IEEE Transactions on Multimedia}}  \bibinfo{volume}{23} (\bibinfo{year}{2020}), \bibinfo{pages}{3278--3291}.
\newblock


\bibitem[Liu et~al\mbox{.}(2021b)]%
        {liu2021pqa}
\bibfield{author}{\bibinfo{person}{Qi Liu}, \bibinfo{person}{Hui Yuan}, \bibinfo{person}{Honglei Su}, \bibinfo{person}{Hao Liu}, \bibinfo{person}{Yu Wang}, \bibinfo{person}{Huan Yang}, {and} \bibinfo{person}{Junhui Hou}.} \bibinfo{year}{2021}\natexlab{b}.
\newblock \showarticletitle{PQA-Net: Deep No Reference Point Cloud Quality Assessment via Multi-View Projection}.
\newblock \bibinfo{journal}{\emph{IEEE Transactions on Circuits and Systems for Video Technology}} \bibinfo{volume}{31}, \bibinfo{number}{12} (\bibinfo{year}{2021}), \bibinfo{pages}{4645--4660}.
\newblock


\bibitem[Liu et~al\mbox{.}(2015)]%
        {liu2015paraboost}
\bibfield{author}{\bibinfo{person}{Tsung-Jung Liu}, \bibinfo{person}{Kuan-Hsien Liu}, \bibinfo{person}{Joe~Yuchieh Lin}, \bibinfo{person}{Weisi Lin}, {and} \bibinfo{person}{C-C~Jay Kuo}.} \bibinfo{year}{2015}\natexlab{}.
\newblock \showarticletitle{A paraboost method to image quality assessment}.
\newblock \bibinfo{journal}{\emph{IEEE Transactions on Neural Networks and Learning Systems}} \bibinfo{volume}{28}, \bibinfo{number}{1} (\bibinfo{year}{2015}), \bibinfo{pages}{107--121}.
\newblock


\bibitem[Liu et~al\mbox{.}(2022a)]%
        {liutoposeg}
\bibfield{author}{\bibinfo{person}{Weiquan Liu}, \bibinfo{person}{Hanyun Guo}, \bibinfo{person}{Weini Zhang}, \bibinfo{person}{Yu Zang}, \bibinfo{person}{Cheng Wang}, {and} \bibinfo{person}{Jonathan Li}.} \bibinfo{year}{2022}\natexlab{a}.
\newblock \showarticletitle{TopoSeg: Topology-aware Segmentation for Point Clouds}.
\newblock \bibinfo{journal}{\emph{International Joint Conference on Artificial Intelligence}} (\bibinfo{year}{2022}).
\newblock


\bibitem[Liu et~al\mbox{.}(2022c)]%
        {liu2022point}
\bibfield{author}{\bibinfo{person}{Yipeng Liu}, \bibinfo{person}{Qi Yang}, \bibinfo{person}{Yiling Xu}, {and} \bibinfo{person}{Le Yang}.} \bibinfo{year}{2022}\natexlab{c}.
\newblock \showarticletitle{Point Cloud Quality Assessment: Dataset Construction and Learning-based No-Reference Metric}.
\newblock \bibinfo{journal}{\emph{ACM Transactions on Multimedia Computing, Communications, and Applications}} (\bibinfo{year}{2022}).
\newblock


\bibitem[Mekuria et~al\mbox{.}(2016a)]%
        {mekuria2016design}
\bibfield{author}{\bibinfo{person}{Rufael Mekuria}, \bibinfo{person}{Kees Blom}, {and} \bibinfo{person}{Pablo Cesar}.} \bibinfo{year}{2016}\natexlab{a}.
\newblock \showarticletitle{Design, implementation, and evaluation of a point cloud codec for tele-immersive video}.
\newblock \bibinfo{journal}{\emph{IEEE Transactions on Circuits and Systems for Video Technology}} \bibinfo{volume}{27}, \bibinfo{number}{4} (\bibinfo{year}{2016}), \bibinfo{pages}{828--842}.
\newblock


\bibitem[Mekuria et~al\mbox{.}(2016b)]%
        {mekuria2016evaluation}
\bibfield{author}{\bibinfo{person}{R Mekuria}, \bibinfo{person}{Z Li}, \bibinfo{person}{C Tulvan}, {and} \bibinfo{person}{P Chou}.} \bibinfo{year}{2016}\natexlab{b}.
\newblock \showarticletitle{Evaluation criteria for point cloud compression}.
\newblock \bibinfo{journal}{\emph{ISO/IEC MPEG}} \bibinfo{number}{16332} (\bibinfo{year}{2016}).
\newblock


\bibitem[Meynet et~al\mbox{.}(2020)]%
        {meynet2020pcqm}
\bibfield{author}{\bibinfo{person}{Gabriel Meynet}, \bibinfo{person}{Yana Nehm{\'e}}, \bibinfo{person}{Julie Digne}, {and} \bibinfo{person}{Guillaume Lavou{\'e}}.} \bibinfo{year}{2020}\natexlab{}.
\newblock \showarticletitle{PCQM: A full-reference quality metric for colored 3D point clouds}. In \bibinfo{booktitle}{\emph{International Workshop on Quality of Multimedia}}. \bibinfo{pages}{1--6}.
\newblock


\bibitem[Mittal et~al\mbox{.}(2012a)]%
        {mittal2012brisque}
\bibfield{author}{\bibinfo{person}{Anish Mittal}, \bibinfo{person}{Anush~Krishna Moorthy}, {and} \bibinfo{person}{Alan~Conrad Bovik}.} \bibinfo{year}{2012}\natexlab{a}.
\newblock \showarticletitle{No-reference image quality assessment in the spatial domain}.
\newblock \bibinfo{journal}{\emph{IEEE Transactions on Image Processing}} \bibinfo{volume}{21}, \bibinfo{number}{12} (\bibinfo{year}{2012}), \bibinfo{pages}{4695--4708}.
\newblock


\bibitem[Mittal et~al\mbox{.}(2012b)]%
        {mittal2012making}
\bibfield{author}{\bibinfo{person}{Anish Mittal}, \bibinfo{person}{Rajiv Soundararajan}, {and} \bibinfo{person}{Alan~C Bovik}.} \bibinfo{year}{2012}\natexlab{b}.
\newblock \showarticletitle{Making a “completely blind” image quality analyzer}.
\newblock \bibinfo{journal}{\emph{IEEE Signal Processing Letters}} \bibinfo{volume}{20}, \bibinfo{number}{3} (\bibinfo{year}{2012}), \bibinfo{pages}{209--212}.
\newblock


\bibitem[Park et~al\mbox{.}(2008)]%
        {park2008multiple}
\bibfield{author}{\bibinfo{person}{Youngmin Park}, \bibinfo{person}{Vincent Lepetit}, {and} \bibinfo{person}{Woontack Woo}.} \bibinfo{year}{2008}\natexlab{}.
\newblock \showarticletitle{Multiple 3d object tracking for augmented reality}. In \bibinfo{booktitle}{\emph{IEEE/ACM International Symposium on Mixed and Augmented Reality}}. \bibinfo{pages}{117--120}.
\newblock


\bibitem[Radford et~al\mbox{.}(2021)]%
        {clip}
\bibfield{author}{\bibinfo{person}{Alec Radford}, \bibinfo{person}{Jong~Wook Kim}, \bibinfo{person}{Chris Hallacy}, \bibinfo{person}{Aditya Ramesh}, \bibinfo{person}{Gabriel Goh}, \bibinfo{person}{Sandhini Agarwal}, \bibinfo{person}{Girish Sastry}, \bibinfo{person}{Amanda Askell}, \bibinfo{person}{Pamela Mishkin}, \bibinfo{person}{Jack Clark}, \bibinfo{person}{Gretchen Krueger}, {and} \bibinfo{person}{Ilya Sutskever}.} \bibinfo{year}{2021}\natexlab{}.
\newblock \bibinfo{title}{Learning Transferable Visual Models From Natural Language Supervision}.
\newblock
\newblock


\bibitem[Radford et~al\mbox{.}(2019)]%
        {gpt2}
\bibfield{author}{\bibinfo{person}{Alec Radford}, \bibinfo{person}{Jeff Wu}, \bibinfo{person}{Rewon Child}, \bibinfo{person}{D. Luan}, \bibinfo{person}{Dario Amodei}, {and} \bibinfo{person}{Ilya Sutskever}.} \bibinfo{year}{2019}\natexlab{}.
\newblock \bibinfo{title}{Language models are unsupervised multitask learners}.
\newblock
\newblock


\bibitem[Tian et~al\mbox{.}(2017)]%
        {tian2017geometric}
\bibfield{author}{\bibinfo{person}{Dong Tian}, \bibinfo{person}{Hideaki Ochimizu}, \bibinfo{person}{Chen Feng}, \bibinfo{person}{Robert Cohen}, {and} \bibinfo{person}{Anthony Vetro}.} \bibinfo{year}{2017}\natexlab{}.
\newblock \showarticletitle{Geometric distortion metrics for point cloud compression}. In \bibinfo{booktitle}{\emph{IEEE International Conference on Image Processing}}. \bibinfo{pages}{3460--3464}.
\newblock


\bibitem[Torlig et~al\mbox{.}(2018)]%
        {torlig2018novel}
\bibfield{author}{\bibinfo{person}{Eric~M Torlig}, \bibinfo{person}{Evangelos Alexiou}, \bibinfo{person}{Tiago~A Fonseca}, \bibinfo{person}{Ricardo~L de Queiroz}, {and} \bibinfo{person}{Touradj Ebrahimi}.} \bibinfo{year}{2018}\natexlab{}.
\newblock \showarticletitle{A novel methodology for quality assessment of voxelized point clouds}. In \bibinfo{booktitle}{\emph{Applications of Digital Image Processing XLI}}, Vol.~\bibinfo{volume}{10752}. \bibinfo{pages}{174--190}.
\newblock


\bibitem[Touvron et~al\mbox{.}(2023)]%
        {llama2}
\bibfield{author}{\bibinfo{person}{Hugo Touvron}, \bibinfo{person}{Louis Martin}, \bibinfo{person}{Kevin Stone}, \bibinfo{person}{Peter Albert}, \bibinfo{person}{Amjad Almahairi}, \bibinfo{person}{Yasmine Babaei}, \bibinfo{person}{Nikolay Bashlykov}, \bibinfo{person}{Soumya Batra}, \bibinfo{person}{Prajjwal Bhargava}, \bibinfo{person}{Shruti Bhosale}, \bibinfo{person}{Dan Bikel}, \bibinfo{person}{Lukas Blecher}, \bibinfo{person}{Cristian~Canton Ferrer}, \bibinfo{person}{Moya Chen}, \bibinfo{person}{Guillem Cucurull}, \bibinfo{person}{David Esiobu}, \bibinfo{person}{Jude Fernandes}, \bibinfo{person}{Jeremy Fu}, \bibinfo{person}{Wenyin Fu}, \bibinfo{person}{Brian Fuller}, \bibinfo{person}{Cynthia Gao}, \bibinfo{person}{Vedanuj Goswami}, \bibinfo{person}{Naman Goyal}, \bibinfo{person}{Anthony Hartshorn}, \bibinfo{person}{Saghar Hosseini}, \bibinfo{person}{Rui Hou}, \bibinfo{person}{Hakan Inan}, \bibinfo{person}{Marcin Kardas}, \bibinfo{person}{Viktor Kerkez}, \bibinfo{person}{Madian Khabsa},
  \bibinfo{person}{Isabel Kloumann}, \bibinfo{person}{Artem Korenev}, \bibinfo{person}{Punit~Singh Koura}, \bibinfo{person}{Marie-Anne Lachaux}, \bibinfo{person}{Thibaut Lavril}, \bibinfo{person}{Jenya Lee}, \bibinfo{person}{Diana Liskovich}, \bibinfo{person}{Yinghai Lu}, \bibinfo{person}{Yuning Mao}, \bibinfo{person}{Xavier Martinet}, \bibinfo{person}{Todor Mihaylov}, \bibinfo{person}{Pushkar Mishra}, \bibinfo{person}{Igor Molybog}, \bibinfo{person}{Yixin Nie}, \bibinfo{person}{Andrew Poulton}, \bibinfo{person}{Jeremy Reizenstein}, \bibinfo{person}{Rashi Rungta}, \bibinfo{person}{Kalyan Saladi}, \bibinfo{person}{Alan Schelten}, \bibinfo{person}{Ruan Silva}, \bibinfo{person}{Eric~Michael Smith}, \bibinfo{person}{Ranjan Subramanian}, \bibinfo{person}{Xiaoqing~Ellen Tan}, \bibinfo{person}{Binh Tang}, \bibinfo{person}{Ross Taylor}, \bibinfo{person}{Adina Williams}, \bibinfo{person}{Jian~Xiang Kuan}, \bibinfo{person}{Puxin Xu}, \bibinfo{person}{Zheng Yan}, \bibinfo{person}{Iliyan Zarov}, \bibinfo{person}{Yuchen
  Zhang}, \bibinfo{person}{Angela Fan}, \bibinfo{person}{Melanie Kambadur}, \bibinfo{person}{Sharan Narang}, \bibinfo{person}{Aurelien Rodriguez}, \bibinfo{person}{Robert Stojnic}, \bibinfo{person}{Sergey Edunov}, {and} \bibinfo{person}{Thomas Scialom}.} \bibinfo{year}{2023}\natexlab{}.
\newblock \bibinfo{title}{Llama 2: Open Foundation and Fine-Tuned Chat Models}.
\newblock
\newblock
\showeprint[arxiv]{2307.09288}~[cs.CL]


\bibitem[Vora et~al\mbox{.}(2020)]%
        {vora2020pointpainting}
\bibfield{author}{\bibinfo{person}{Sourabh Vora}, \bibinfo{person}{Alex~H Lang}, \bibinfo{person}{Bassam Helou}, {and} \bibinfo{person}{Oscar Beijbom}.} \bibinfo{year}{2020}\natexlab{}.
\newblock \showarticletitle{Pointpainting: Sequential fusion for 3d object detection}. In \bibinfo{booktitle}{\emph{Proceedings of the IEEE/CVF international conference on computer vision}}. \bibinfo{pages}{4604--4612}.
\newblock


\bibitem[Wang et~al\mbox{.}(2023)]%
        {wang2023non}
\bibfield{author}{\bibinfo{person}{Songtao Wang}, \bibinfo{person}{Xiaoqi Wang}, \bibinfo{person}{Hao Gao}, {and} \bibinfo{person}{Jian Xiong}.} \bibinfo{year}{2023}\natexlab{}.
\newblock \showarticletitle{Non-Local Geometry and Color Gradient Aggregation Graph Model for No-Reference Point Cloud Quality Assessment}. In \bibinfo{booktitle}{\emph{Proceedings of the 31st ACM International Conference on Multimedia}}. \bibinfo{pages}{6803--6810}.
\newblock


\bibitem[Wang and Jia(2019)]%
        {wang2019frustum}
\bibfield{author}{\bibinfo{person}{Zhixin Wang} {and} \bibinfo{person}{Kui Jia}.} \bibinfo{year}{2019}\natexlab{}.
\newblock \showarticletitle{Frustum convnet: Sliding frustums to aggregate local point-wise features for amodal 3d object detection}. In \bibinfo{booktitle}{\emph{IEEE/RSJ International Conference on Intelligent Robots and Systems}}. \bibinfo{pages}{1742--1749}.
\newblock


\bibitem[Wu et~al\mbox{.}(2024a)]%
        {wu2023qbench}
\bibfield{author}{\bibinfo{person}{Haoning Wu}, \bibinfo{person}{Zicheng Zhang}, \bibinfo{person}{Erli Zhang}, \bibinfo{person}{Chaofeng Chen}, \bibinfo{person}{Liang Liao}, \bibinfo{person}{Annan Wang}, \bibinfo{person}{Chunyi Li}, \bibinfo{person}{Wenxiu Sun}, \bibinfo{person}{Qiong Yan}, \bibinfo{person}{Guangtao Zhai}, {and} \bibinfo{person}{Weisi Lin}.} \bibinfo{year}{2024}\natexlab{a}.
\newblock \showarticletitle{Q-Bench: A Benchmark for General-Purpose Foundation Models on Low-level Vision}.
\newblock \bibinfo{journal}{\emph{ICLR}} (\bibinfo{year}{2024}).
\newblock


\bibitem[Wu et~al\mbox{.}(2024b)]%
        {wu2023qinstruct}
\bibfield{author}{\bibinfo{person}{Haoning Wu}, \bibinfo{person}{Zicheng Zhang}, \bibinfo{person}{Erli Zhang}, \bibinfo{person}{Chaofeng Chen}, \bibinfo{person}{Liang Liao}, \bibinfo{person}{Annan Wang}, \bibinfo{person}{Kaixin Xu}, \bibinfo{person}{Chunyi Li}, \bibinfo{person}{Jingwen Hou}, \bibinfo{person}{Guangtao Zhai}, {et~al\mbox{.}}} \bibinfo{year}{2024}\natexlab{b}.
\newblock \showarticletitle{Q-instruct: Improving low-level visual abilities for multi-modality foundation models}.
\newblock \bibinfo{journal}{\emph{CVPR}} (\bibinfo{year}{2024}).
\newblock


\bibitem[Wu et~al\mbox{.}(2023)]%
        {wu2023qalign}
\bibfield{author}{\bibinfo{person}{Haoning Wu}, \bibinfo{person}{Zicheng Zhang}, \bibinfo{person}{Weixia Zhang}, \bibinfo{person}{Chaofeng Chen}, \bibinfo{person}{Liang Liao}, \bibinfo{person}{Chunyi Li}, \bibinfo{person}{Yixuan Gao}, \bibinfo{person}{Annan Wang}, \bibinfo{person}{Erli Zhang}, \bibinfo{person}{Wenxiu Sun}, {et~al\mbox{.}}} \bibinfo{year}{2023}\natexlab{}.
\newblock \showarticletitle{Q-align: Teaching lmms for visual scoring via discrete text-defined levels}.
\newblock \bibinfo{journal}{\emph{arXiv preprint arXiv:2312.17090}} (\bibinfo{year}{2023}).
\newblock


\bibitem[Wu et~al\mbox{.}(2024c)]%
        {coinstruct}
\bibfield{author}{\bibinfo{person}{Haoning Wu}, \bibinfo{person}{Hanwei Zhu}, \bibinfo{person}{Zicheng Zhang}, \bibinfo{person}{Erli Zhang}, \bibinfo{person}{Chaofeng Chen}, \bibinfo{person}{Liang Liao}, \bibinfo{person}{Chunyi Li}, \bibinfo{person}{Annan Wang}, \bibinfo{person}{Wenxiu Sun}, \bibinfo{person}{Qiong Yan}, \bibinfo{person}{Xiaohong Liu}, \bibinfo{person}{Guangtao Zhai}, \bibinfo{person}{Shiqi Wang}, {and} \bibinfo{person}{Weisi Lin}.} \bibinfo{year}{2024}\natexlab{c}.
\newblock \showarticletitle{Towards Open-ended Visual Quality Comparison}.
\newblock \bibinfo{journal}{\emph{arXiv preprint arXiv:2402.16641}} (\bibinfo{year}{2024}).
\newblock


\bibitem[Xie et~al\mbox{.}(2020)]%
        {xie2020pi}
\bibfield{author}{\bibinfo{person}{Liang Xie}, \bibinfo{person}{Chao Xiang}, \bibinfo{person}{Zhengxu Yu}, \bibinfo{person}{Guodong Xu}, \bibinfo{person}{Zheng Yang}, \bibinfo{person}{Deng Cai}, {and} \bibinfo{person}{Xiaofei He}.} \bibinfo{year}{2020}\natexlab{}.
\newblock \showarticletitle{PI-RCNN: An efficient multi-sensor 3D object detector with point-based attentive cont-conv fusion module}. In \bibinfo{booktitle}{\emph{AAAI}}, Vol.~\bibinfo{volume}{34}. \bibinfo{pages}{12460--12467}.
\newblock


\bibitem[Xie et~al\mbox{.}(2023)]%
        {xie2023pmbqa}
\bibfield{author}{\bibinfo{person}{Wuyuan Xie}, \bibinfo{person}{Kaimin Wang}, \bibinfo{person}{Yakun Ju}, {and} \bibinfo{person}{Miaohui Wang}.} \bibinfo{year}{2023}\natexlab{}.
\newblock \showarticletitle{pmbqa: Projection-based blind point cloud quality assessment via multimodal learning}. In \bibinfo{booktitle}{\emph{Proceedings of the 31st ACM International Conference on Multimedia}}. \bibinfo{pages}{3250--3258}.
\newblock


\bibitem[Yang et~al\mbox{.}(2020a)]%
        {yang2020predicting}
\bibfield{author}{\bibinfo{person}{Qi Yang}, \bibinfo{person}{Hao Chen}, \bibinfo{person}{Zhan Ma}, \bibinfo{person}{Yiling Xu}, \bibinfo{person}{Rongjun Tang}, {and} \bibinfo{person}{Jun Sun}.} \bibinfo{year}{2020}\natexlab{a}.
\newblock \showarticletitle{Predicting the perceptual quality of point cloud: A 3d-to-2d projection-based exploration}.
\newblock \bibinfo{journal}{\emph{IEEE Transactions on Multimedia}} (\bibinfo{year}{2020}).
\newblock


\bibitem[Yang et~al\mbox{.}(2022)]%
        {yang2022no}
\bibfield{author}{\bibinfo{person}{Qi Yang}, \bibinfo{person}{Yipeng Liu}, \bibinfo{person}{Siheng Chen}, \bibinfo{person}{Yiling Xu}, {and} \bibinfo{person}{Jun Sun}.} \bibinfo{year}{2022}\natexlab{}.
\newblock \showarticletitle{No-Reference Point Cloud Quality Assessment via Domain Adaptation}. In \bibinfo{booktitle}{\emph{Proceedings of the IEEE/CVF international conference on computer vision}}. \bibinfo{pages}{21179--21188}.
\newblock


\bibitem[Yang et~al\mbox{.}(2020b)]%
        {yang2020graphsim}
\bibfield{author}{\bibinfo{person}{Qi Yang}, \bibinfo{person}{Zhan Ma}, \bibinfo{person}{Yiling Xu}, \bibinfo{person}{Zhu Li}, {and} \bibinfo{person}{Jun Sun}.} \bibinfo{year}{2020}\natexlab{b}.
\newblock \showarticletitle{Inferring point cloud quality via graph similarity}.
\newblock \bibinfo{journal}{\emph{IEEE Transactions on Pattern Analysis and Machine Intelligence}} (\bibinfo{year}{2020}).
\newblock


\bibitem[Ye et~al\mbox{.}(2023a)]%
        {mplug}
\bibfield{author}{\bibinfo{person}{Qinghao Ye}, \bibinfo{person}{Haiyang Xu}, \bibinfo{person}{Guohai Xu}, \bibinfo{person}{Jiabo Ye}, \bibinfo{person}{Ming Yan}, \bibinfo{person}{Yiyang Zhou}, \bibinfo{person}{Junyang Wang}, \bibinfo{person}{Anwen Hu}, \bibinfo{person}{Pengcheng Shi}, \bibinfo{person}{Yaya Shi}, \bibinfo{person}{Chaoya Jiang}, \bibinfo{person}{Chenliang Li}, \bibinfo{person}{Yuanhong Xu}, \bibinfo{person}{Hehong Chen}, \bibinfo{person}{Junfeng Tian}, \bibinfo{person}{Qian Qi}, \bibinfo{person}{Ji Zhang}, {and} \bibinfo{person}{Fei Huang}.} \bibinfo{year}{2023}\natexlab{a}.
\newblock \bibinfo{title}{mPLUG-Owl: Modularization Empowers Large Language Models with Multimodality}.
\newblock
\newblock
\showeprint[arxiv]{2304.14178}~[cs.CL]


\bibitem[Ye et~al\mbox{.}(2023b)]%
        {mplugowl2}
\bibfield{author}{\bibinfo{person}{Qinghao Ye}, \bibinfo{person}{Haiyang Xu}, \bibinfo{person}{Jiabo Ye}, \bibinfo{person}{Ming Yan}, \bibinfo{person}{Anwen Hu}, \bibinfo{person}{Haowei Liu}, \bibinfo{person}{Qi Qian}, \bibinfo{person}{Ji Zhang}, \bibinfo{person}{Fei Huang}, {and} \bibinfo{person}{Jingren Zhou}.} \bibinfo{year}{2023}\natexlab{b}.
\newblock \bibinfo{title}{mPLUG-Owl2: Revolutionizing Multi-modal Large Language Model with Modality Collaboration}.
\newblock
\newblock
\showeprint[arxiv]{2311.04257}~[cs.CL]


\bibitem[Yoo et~al\mbox{.}(2020)]%
        {yoo20203d}
\bibfield{author}{\bibinfo{person}{Jin~Hyeok Yoo}, \bibinfo{person}{Yecheol Kim}, \bibinfo{person}{Jisong Kim}, {and} \bibinfo{person}{Jun~Won Choi}.} \bibinfo{year}{2020}\natexlab{}.
\newblock \showarticletitle{3d-cvf: Generating joint camera and lidar features using cross-view spatial feature fusion for 3d object detection}. In \bibinfo{booktitle}{\emph{European Conference on Computer Vision}}. \bibinfo{pages}{720--736}.
\newblock


\bibitem[You et~al\mbox{.}(2023)]%
        {depictqa}
\bibfield{author}{\bibinfo{person}{Zhiyuan You}, \bibinfo{person}{Zheyuan Li}, \bibinfo{person}{Jinjin Gu}, \bibinfo{person}{Zhenfei Yin}, \bibinfo{person}{Tianfan Xue}, {and} \bibinfo{person}{Chao Dong}.} \bibinfo{year}{2023}\natexlab{}.
\newblock \bibinfo{title}{Depicting Beyond Scores: Advancing Image Quality Assessment through Multi-modal Language Models}.
\newblock
\newblock
\showeprint[arxiv]{2312.08962}~[cs.CV]


\bibitem[Zhang et~al\mbox{.}(2022a)]%
        {zhang2022dual}
\bibfield{author}{\bibinfo{person}{Chaofan Zhang}, \bibinfo{person}{Ziqing Huang}, \bibinfo{person}{Shiguang Liu}, {and} \bibinfo{person}{Jian Xiao}.} \bibinfo{year}{2022}\natexlab{a}.
\newblock \showarticletitle{Dual-Channel Multi-Task CNN for No-Reference Screen Content Image Quality Assessment}.
\newblock \bibinfo{journal}{\emph{IEEE Transactions on Circuits and Systems for Video Technology}} \bibinfo{volume}{32}, \bibinfo{number}{8} (\bibinfo{year}{2022}), \bibinfo{pages}{5011--5025}.
\newblock


\bibitem[Zhang and Liu(2022)]%
        {zhang2022nor}
\bibfield{author}{\bibinfo{person}{Chaofan Zhang} {and} \bibinfo{person}{Shiguang Liu}.} \bibinfo{year}{2022}\natexlab{}.
\newblock \showarticletitle{No-reference omnidirectional image quality assessment based on joint network}. In \bibinfo{booktitle}{\emph{ACM International Conference on Multimedia}}. \bibinfo{pages}{943--951}.
\newblock


\bibitem[Zhang et~al\mbox{.}(2015b)]%
        {zhang2015feature}
\bibfield{author}{\bibinfo{person}{Lin Zhang}, \bibinfo{person}{Lei Zhang}, {and} \bibinfo{person}{Alan~C Bovik}.} \bibinfo{year}{2015}\natexlab{b}.
\newblock \showarticletitle{A feature-enriched completely blind image quality evaluator}.
\newblock \bibinfo{journal}{\emph{IEEE Transactions on Image Processing}} \bibinfo{volume}{24}, \bibinfo{number}{8} (\bibinfo{year}{2015}), \bibinfo{pages}{2579--2591}.
\newblock


\bibitem[Zhang et~al\mbox{.}(2015a)]%
        {zhang2015application}
\bibfield{author}{\bibinfo{person}{Wei Zhang}, \bibinfo{person}{Ali Borji}, \bibinfo{person}{Zhou Wang}, \bibinfo{person}{Patrick Le~Callet}, {and} \bibinfo{person}{Hantao Liu}.} \bibinfo{year}{2015}\natexlab{a}.
\newblock \showarticletitle{The application of visual saliency models in objective image quality assessment: A statistical evaluation}.
\newblock \bibinfo{journal}{\emph{IEEE Transactions on Neural Networks and Learning Systems}} \bibinfo{volume}{27}, \bibinfo{number}{6} (\bibinfo{year}{2015}), \bibinfo{pages}{1266--1278}.
\newblock


\bibitem[Zhang et~al\mbox{.}(2021b)]%
        {zhang2021ms}
\bibfield{author}{\bibinfo{person}{Yujie Zhang}, \bibinfo{person}{Qi Yang}, {and} \bibinfo{person}{Yiling Xu}.} \bibinfo{year}{2021}\natexlab{b}.
\newblock \showarticletitle{MS-GraphSIM: Inferring point cloud quality via multiscale graph similarity}. In \bibinfo{booktitle}{\emph{Proceedings of the 29th ACM International Conference on Multimedia}}. \bibinfo{pages}{1230--1238}.
\newblock


\bibitem[Zhang et~al\mbox{.}(2022b)]%
        {zhang2021no}
\bibfield{author}{\bibinfo{person}{Zicheng Zhang}, \bibinfo{person}{Wei Sun}, \bibinfo{person}{Xiongkuo Min}, \bibinfo{person}{Tao Wang}, \bibinfo{person}{Wei Lu}, {and} \bibinfo{person}{Guangtao Zhai}.} \bibinfo{year}{2022}\natexlab{b}.
\newblock \showarticletitle{No-reference quality assessment for 3d colored point cloud and mesh models}.
\newblock \bibinfo{journal}{\emph{IEEE Transactions on Circuits and Systems for Video Technology}} (\bibinfo{year}{2022}).
\newblock


\bibitem[Zhang et~al\mbox{.}(2022c)]%
        {zhang2022treating}
\bibfield{author}{\bibinfo{person}{Zicheng Zhang}, \bibinfo{person}{Wei Sun}, \bibinfo{person}{Xiongkuo Min}, \bibinfo{person}{Wei Wu}, \bibinfo{person}{Ying Chen}, {and} \bibinfo{person}{Guangtao Zhai}.} \bibinfo{year}{2022}\natexlab{c}.
\newblock \showarticletitle{Treating Point Cloud as Moving Camera Videos: A No-Reference Quality Assessment Metric}.
\newblock \bibinfo{journal}{\emph{arXiv preprint arXiv:2208.14085}} (\bibinfo{year}{2022}).
\newblock


\bibitem[Zhang et~al\mbox{.}(2023a)]%
        {zhang2022mm}
\bibfield{author}{\bibinfo{person}{Zicheng Zhang}, \bibinfo{person}{Wei Sun}, \bibinfo{person}{Xiongkuo Min}, \bibinfo{person}{Quan Zhou}, \bibinfo{person}{Jun He}, \bibinfo{person}{Qiyuan Wang}, {and} \bibinfo{person}{Guangtao Zhai}.} \bibinfo{year}{2023}\natexlab{a}.
\newblock \showarticletitle{MM-PCQA: Multi-modal learning for no-reference point cloud quality assessment}.
\newblock \bibinfo{journal}{\emph{International Joint Conference on Artificial Intelligence}} (\bibinfo{year}{2023}).
\newblock


\bibitem[Zhang et~al\mbox{.}(2021a)]%
        {zhang2021ano}
\bibfield{author}{\bibinfo{person}{Zicheng Zhang}, \bibinfo{person}{Wei Sun}, \bibinfo{person}{Xiongkuo Min}, \bibinfo{person}{Wenhan Zhu}, \bibinfo{person}{Tao Wang}, \bibinfo{person}{Wei Lu}, {and} \bibinfo{person}{Guangtao Zhai}.} \bibinfo{year}{2021}\natexlab{a}.
\newblock \showarticletitle{A No-Reference Evaluation Metric for Low-Light Image Enhancement}. In \bibinfo{booktitle}{\emph{IEEE International Conference on Multimedia and Expo}}.
\newblock


\bibitem[Zhang et~al\mbox{.}(2023b)]%
        {zhang2023gms}
\bibfield{author}{\bibinfo{person}{Zicheng Zhang}, \bibinfo{person}{Wei Sun}, \bibinfo{person}{Houning Wu}, \bibinfo{person}{Yingjie Zhou}, \bibinfo{person}{Chunyi Li}, \bibinfo{person}{Xiongkuo Min}, \bibinfo{person}{Guangtao Zhai}, {and} \bibinfo{person}{Weisi Lin}.} \bibinfo{year}{2023}\natexlab{b}.
\newblock \showarticletitle{GMS-3DQA: Projection-based Grid Mini-patch Sampling for 3D Model Quality Assessment}.
\newblock \bibinfo{journal}{\emph{arXiv preprint arXiv:2306.05658}} (\bibinfo{year}{2023}).
\newblock


\bibitem[Zhou et~al\mbox{.}(2018)]%
        {zhou2018open3d}
\bibfield{author}{\bibinfo{person}{Qian-Yi Zhou}, \bibinfo{person}{Jaesik Park}, {and} \bibinfo{person}{Vladlen Koltun}.} \bibinfo{year}{2018}\natexlab{}.
\newblock \showarticletitle{Open3D: A modern library for 3D data processing}.
\newblock \bibinfo{journal}{\emph{arXiv preprint arXiv:1801.09847}} (\bibinfo{year}{2018}).
\newblock


\bibitem[Zhou et~al\mbox{.}(2022)]%
        {zhou2022blind}
\bibfield{author}{\bibinfo{person}{Wei Zhou}, \bibinfo{person}{Qi Yang}, \bibinfo{person}{Qiuping Jiang}, \bibinfo{person}{Guangtao Zhai}, {and} \bibinfo{person}{Weisi Lin}.} \bibinfo{year}{2022}\natexlab{}.
\newblock \showarticletitle{Blind Quality Assessment of 3D Dense Point Clouds with Structure Guided Resampling}.
\newblock \bibinfo{journal}{\emph{arXiv preprint arXiv:2208.14603}} (\bibinfo{year}{2022}).
\newblock


\bibitem[Zhou et~al\mbox{.}(2023)]%
        {zhou2023pointpca+}
\bibfield{author}{\bibinfo{person}{Xuemei Zhou}, \bibinfo{person}{Evangelos Alexiou}, \bibinfo{person}{Irene Viola}, {and} \bibinfo{person}{Pablo Cesar}.} \bibinfo{year}{2023}\natexlab{}.
\newblock \showarticletitle{PointPCA+: Extending PointPCA objective quality assessment metric}. In \bibinfo{booktitle}{\emph{2023 IEEE International Conference on Image Processing Challenges and Workshops (ICIPCW)}}. IEEE, \bibinfo{pages}{1--5}.
\newblock


\end{thebibliography}

\end{document}